\documentclass[letterpaper,10pt]{article} 
\usepackage{osameet3} 
\usepackage[square,numbers]{natbib} 
\usepackage{amsmath,appendix}
\usepackage{url}       
\usepackage{booktabs}       
\usepackage[colorlinks=true,bookmarks=false,citecolor=blue,urlcolor=blue]{hyperref} 
\usepackage{graphicx}
\usepackage[utf8]{inputenc}
\usepackage[linesnumbered,lined,boxed,commentsnumbered,ruled,vlined]{algorithm2e}
\usepackage{dsfont}
\usepackage{amsthm}
\usepackage{stackrel}
\newtheorem{theorem}{Theorem}
\newtheorem{lemma}{Lemma}
\newtheorem{proposition}{Proposition}
\theoremstyle{definition}

\newtheorem{example}{Example}
\newtheorem{assumption}{Assumption}

\newtheorem{corollary}{Corollary}
\theoremstyle{remark}
\newtheorem{remark}{Remark}
\usepackage{hyperref}
\usepackage{cleveref}
\crefname{subsection}{subsection}{subsections}
\crefname{lemma}{lemma}{lemma}
\crefname{property}{property}{property}
\crefname{table}{table}{table}
\crefname{assumption}{assumption}{assumptions}
\usepackage{libertine}
\usepackage[libertine]{newtxmath}
\usepackage{parskip}
\usepackage{notation}
\usepackage[depth=4]{bookmark}

\usepackage{tikz}
\usetikzlibrary{automata,arrows,positioning,calc}

\begin{document}

\title{Finite time analysis of temporal difference learning with linear function approximation: Tail averaging and regularisation\footnotemark}

\author{Gandharv Patil$^{1}$, Prashanth L.A.$^{2}$, Dheeraj Nagaraj$^{3}$, Doina Precup$^{1}$}
\vspace{0.5cm}
\address{1: McGill University \& Mila, 2: Indian Institute of Technology Madras, 3: Google Research}
\vspace{0.5cm}
\email{gandharv.patil@mail.mcgill.ca, prashla@cse.iitm.ac.in, dheerajnagaraj@google.com, dprecup@cs.mcgill.ca}
\footnotetext{This is longer version of the paper presented at AISTATS 2023~\citep{pmlr-v206-patil23a}}
\begin{abstract}
We study the finite-time behaviour of the popular temporal difference (TD) learning algorithm when combined with tail-averaging. We derive finite time bounds on the parameter error of the tail-averaged TD iterate under a step-size choice that does not require information about the eigenvalues of the matrix underlying the projected TD fixed point. Our analysis shows that tail-averaged TD converges at the optimal $O\left(1/t\right)$ rate, both in expectation and with high probability. In addition, our bounds exhibit a sharper rate of decay for the initial error (bias), which is an improvement over averaging all iterates. We also propose and analyse a variant of TD that incorporates regularisation. From analysis, we conclude that the regularised version of TD is useful for problems with ill-conditioned features.
\end{abstract}

\section{Introduction}
\label{sec:intro}
Temporal difference (TD)~\citep{Sutton88} learning is an efficient and easy to implement stochastic approximation algorithm used for evaluating the long-term performance of a decision policy. The algorithm predicts the value function using a single sample path obtained by simulating the Markov decision process (MDP) with a given policy. Analysis of TD algorithms is challenging, and researchers have devoted significant effort in studying its asymptotic properties~\citep{bvrtd,Pineda97,Schapire2004OnTW,JaakkolaJS94}. 
In recent years, there has been an interest in characterising the finite-time behaviour of TD, and several papers~\citep{DalalSTM18,lstd-prashanth,Bhandari0S18,lakshminarayanan18a,devraj} have tackled this problem under various assumptions. 

For $\timestep$ iterations/updates, most existing works either provide a $O\left(\frac{1}{\timestep^\alpha}\right)$ (with universal step-size)~\citep{DalalSTM18,Bhandari0S18} or a $O\left(\frac{1}{\timestep}\right)$ (with constant step-size)~\citep{lstd-prashanth,Bhandari0S18,lakshminarayanan18a} convergence rate to the TD-fixed point $\params^\star$ defined as $\params^\star \define A^{-1}b$, where $A$ and $b$ are quantities which depend on the MDP and the policy (see Section \ref{sec:model} for the notational information). To obtain a $O\left(\frac{1}{\timestep}\right)$ rate with a constant step-size,~\citep{lstd-prashanth,Bhandari0S18} assume that the minimum eigenvalue of the matrix $A$ is known apriori.  However, in a typical RL setting, such eigenvalue information is not available. Estimating the matrix $A$ and its lowest eigenvalue accurately might require a large number of additional samples, which makes the algorithm more complicated. Therefore, obtaining a $O\left(\frac{1}{t}\right)$ rate for TD with a \textit{universal} step-size is an important open problem. 

In this paper, we provide a solution to this problem by establishing a $O\left(\frac{1}{t}\right)$ bound on the convergence rate for a variant of TD that incorporates tail-averaging, and uses a constant `universal' step-size. In~\citep{Bhandari0S18,lstd-prashanth,lakshminarayanan18a} the authors study an alternate version called iterate averaging which was introduced independently by \citet{polyak1992acceleration} and \citet{ruppert1991stochastic} for general stochastic-approximation algorithms. A shortcoming of iterate averaging is that the initialisation error (i.e., distance between $\params_0$ and $\params^\star$) is forgotten at a slower rate than the non-averaged case, and in practical implementations, one usually performs averaging after a sufficient number of iterations have been performed. This type of delayed averaging, called `tail-averaging', has been explored in the context of ordinary least squares by \citeauthor{jain2018parallelizing} in \citep{jain2018parallelizing}.

Inspired by the analysis of TD learning, we propose a variant of TD that incorporates regularisation, wherein we introduce a  parameter $\lambda$ and solve for the regularised TD fixed point given by $\params^{\star}_{\mathsf{reg}} = (A+\lambda \identity)^{-1}b$. The update rule for this algorithm is similar to vanilla TD except that it involves an additional factor with $\lambda$. Through our analysis we observe that using regularisation can be helpful in obtaining better non-asymptotic bounds for many problems, where the discount factor is close to $1$.

Concretely, the contributions of this paper are as follows: First, we establish a $O(1/t)$ finite time bounds on the convergence rate of tail-averaged TD and tail-averaged TD with regularisation. Similar to \citep{Bhandari0S18,DalalSTM18}, the analysis assumes that the data is sampled in an i.i.d. fashion from a fixed distribution. The resulting bounds are valid under a universal step-size and hold in expectation as well as high probability. We also show that Markov sampling can be handled with simple mixing arguments. The salient features of the bounds for each variant are as follows:

{\textit{Tail averaged TD}}: In this variant, the step-size is a function of the discount factor and a bound on the norm of the state features. The expectation bound provides a $O(1/t)$ convergence rate for tail-averaged TD iterate, while the high-probability bound establishes an exponential concentration of tail-averaged TD around the projected TD fixed point.

{\textit{Tail-averaged TD with regularisation}}: For this variant, the step-size is a function of the discount factor, regularisation parameter $\lambda$, and a bound on the norm of the state features. Although this variant converges to the regularised TD fixed-point $\theta^{\star}_{\mathsf{reg}}$, we show that the worse-case bound on the difference between TD fixed point $\theta^{\star}$ and $\theta^{\star}_{\mathsf{reg}}$ is $O(\lambda)$ in the $\ell_2$ norm. Moreover, our analysis makes a case for using the regularised TD algorithm for problems with ill-conditioned features.

Next, we show that under mixing assumptions, we can extend our results to Markov sampling instead of i.i.d. sampling. These error bounds contain an extra $\tilde{O}(\tau_{\mathsf{mix}})$, where $\tau_{\mathsf{mix}}$ is the underlying Markov chain's mixing time. This is no better than making the samples appear approximately i.i.d. by considering one out of every $\tilde{O}(\tau_{\mathsf{mix}})$ samples, and then dropping the rest. In fact, as per~\citet[Theorem 2]{nagaraj2020least}, even with the discount factor $\beta = 0$, it is information-theoretically impossible to do any better without further assumptions on the nature of the linear approximation. Recently \citet{agarwal} showed that for linear MDPs, one can use reverse experience replay with function approximation to obtain finite time bounds which are independent of the mixing time constant. We leave the study of TD with different experience replay strategies as an interesting future direction, and for the sake of completeness, present the bounds for Markov sampling in Remark~\ref{rem:markov}, and provide a proof sketch in \Cref{sec:mixing}.

In \Cref{tab:summary-expec} we compare our expectation bounds with existing bounds in the literature. In addition, we also derive high-probability bounds for tail-averaged TD with/without regularisation, and we provide a summary of these bounds in a tabular form in \Cref{tab:summary-hpb}.

\begin{table}[htb!]
	\caption{Summary of the bounds in expectation of the form $\mathbb{E}[\norm{\params_{\text{Alg}, \timestep} - \params^{\star}}^{2}_{2}]$, 
	where $\params^{\star}$ is the TD fixed point, and $\params_{\text{Alg},\timestep}$ is the parameter picked by an algorithm  after $\timestep$ iterations of TD.}
	\label{tab:summary-expec}
	\centering
	\resizebox{\columnwidth}{!}{
		\begin{tabular}{|p{1.3in}|p{1.04in}|c|c|}
			\toprule
			Reference & Algorithm & Step-size & Rate\\\midrule
			\citet{Bhandari0S18} & Last iterate & $c/t$\textsuperscript{1} & $O(1/t)$\\
			& Averaged iterate & $\frac{1}{\sqrt{t}}$ & $O(1/\sqrt{t})$\\
			\midrule
			\citet{DalalSTM18} & Last iterate & $1/t^\alpha$ & $O(1/t^\alpha)$ \\\midrule
			\citet{lakshminarayanan18a} &Constant step-size with averaging & $c$ &  $O(1/t)$\\
			\midrule
			 \citet{lstd-prashanth} & Last iterate &$c/n$, $c \propto 1/\mu$ & $O(1/t)$\\
			 & Averaged iterate &$c/t^\alpha$, $c>0$ & $O(1/t^{\alpha})$\\\midrule
			\textbf{Our work} & Tail-averaged TD &$c >0$ & $O(1/t)$\\
			 & Regularised TD\textsuperscript{2} & $c>0$ & $O(1/t)$\\
			 \bottomrule
			 \multicolumn{4}{l}{\textsuperscript{1}\footnotesize{Step-size requires information about eigenvalue of the feature covariance matrix $\Sigma$.}}\\
			 \multicolumn{4}{l}{\textsuperscript{2}\footnotesize{The convergence here is to the regularised TD solution.}}\\
		\end{tabular}
	}
\end{table}

\begin{table}[htb!]
	\caption{Summary of the high-probability bounds of the form $\mathbb{P}\left[\norm{\params_{\text{Alg}, \timestep} - \params^{\star}}^{2}_{2} \le h(t) \right]$, 
	 where $\params^{\star}$ is the TD fixed point, $\params_{\text{Alg},\timestep}$ is the parameter picked by an algorithm  after $\timestep$ iterations of TD, and $h(t)$ is a function of $t$ that depends on $\text{Alg}$.}
	\label{tab:summary-hpb}
	\centering
	\resizebox{\columnwidth}{!}{
		\begin{tabular}{|c|c|c|c|}
			\toprule
			Reference & Algorithm & Step-size & $h(t)$\\\midrule
			\citet{DalalSTM18} & Last iterate & $1/t^\alpha$ & $O(1/t^\alpha)$ \\\midrule
			 \citet{lstd-prashanth} & Last iterate &$c/n$, $c \propto 1/\mu$ & $O(1/t)$\\
			 & Averaged iterate &$c/t^\alpha$, $c>0$ & $O(1/t^{\alpha})$\\\midrule
			\textbf{Our work} & Tail-averaged TD &$c>0$ & $O(1/t)$\\
			 & Regularised TD\textsuperscript{2} & $c>0$ & $O(1/t)$\\
			 \bottomrule
			 \multicolumn{4}{l}{\textsuperscript{2}\footnotesize{The convergence here is to the regularised TD solution.}}
		\end{tabular}
	}
\end{table}

\paragraph{Related work.} Over the past few years, there has been significant interest in understanding the finite-time behaviour of TD learning. Several researchers have proposed interesting frameworks establishing bounds on TD's convergence rate under different assumptions. In \citep{srikant19a,avg-td-Zhang2021FiniteSA,Durmus2021OnTS,WangG20a,Mkvjump} the authors analyse the finite time behaviour of TD using Lyapunov drift-conditions and establish finite time bounds that hold under expectation. The advantage of this framework is that it can be used directly for analysing TD with Markov noise. However, to provide an $O(1/t)$ bound, these analyses use a step-size which depends on the eigenvalue of $A$. For~\eg in \citep[Theorem 7]{srikant19a}, we have $\epsilon = O(\frac{\log T}{\gamma_{\max}T})$ where $\gamma_{\max}$ is essentially the smallest eigenvalue of $A$. Similar conditions can also be found in \citep[Eq. (88)]{Durmus2021OnTS}, \citep[Proposition 2]{WangG20a}, and \citep[Eq. (18)]{Mkvjump}.   

The analysis presented in this work is closely related to bounds established in \citep{Bhandari0S18,lstd-prashanth,lakshminarayanan18a}, where the authors provide an $O(1/t)$ bound in expectation on the mean square error of the parameters. Our bounds match the overall order of these bounds under comparable assumptions. The principal advantage with our bounds is that they hold for a `universal' step-size choice, while the aforementioned references required the knowledge of $\mu$. Another advantage with our bounds, owing to tail averaging, is that the initial error is forgotten exponentially fast, while the corresponding term in the aforementioned references exhibit a power law decay. In another related work, for a universal step-size the authors in \citep{DalalSTM18} provide a $O(1/t^\alpha)$ bound in expectation, where $\alpha \in (0,1)$, while we obtain a $O(1/t)$ bound under similar assumptions. 

Finally, high-probability bounds for TD have been derived in \citep{DalalSTM18,lstd-prashanth}. In comparison to these works, the high-probability bound that we derive is easy to interpret and exhibits better concentration properties. The related Q-learning algorithm and modifications have also been considered in the finite-time regime with linear function approximation (cf. \citep{chen2022finite,chen2022target} and the references therein). However, these results too require the knowledge of the condition number to set the step-size. 

The rest of the paper is organised as follows: In \Cref{sec:model}, we present the main model of TD with function approximation used for our analysis. In \Cref{sec:tailavgtd}, we describe the tail-averaged TD algorithm, and also present the finite time bounds for this algorithm. In \Cref{sec:regtd}, we combine tail-averaging with regularisation in a TD algorithm, and provide finite time bounds for this algorithm. In \Cref{sec:proofsketch}, we present a sketch of the proofs of our main results, and the detailed proofs are available in \cref{sec:convergence-analysis}. In \Cref{sec:mixing}, we discuss the extension of our results to address the case of Markov sampling. Finally, in \Cref{sec:conclusion}, we provide the concluding remarks.


\section{TD with linear function approximation}
\label{sec:model}
Consider an MDP $ \langle \statespace, \actionspace, \prob, \cost, \discount \rangle$, where $\statespace$ is the state space, $\actionspace$ is the action space, $\prob(\state'|\state, \action)$ is the probability of transitioning to the state $\state'$ from the  state $\state$ on choosing action $\action$, $\cost:\statespace \times \actionspace \to \real$ is the per step reward, and $\discount\in [0,1)$ is the discount factor. We assume that the state and action spaces are both finite.
A stationary randomised policy $\policy$ maps every state $\state$ to a distribution over actions. 
For a given policy $\policy$, we define the value function $V^\policy$ as follows:
\begin{align}
    V^{\policy}(\state) = \expecun{}^{\policy, P}\bigg[\sum_{\timestep=0}^{\infty}\discount^{\timestep}\cost(\state_{\timestep},\action_{\timestep})|\State_{0} = \state\bigg], \label{eq:valf}
\end{align}
where the action $\action_t$ in state $\state_t$ is chosen using policy $\policy$, i.e., $\action_\timestep \sim \policy(\state_\timestep)$.
The value function $V^\policy$ obeys the Bellman equation $\bellmanoperator^{\policy}V^{\policy} = V^{\policy}$, where the Bellman operator $\bellmanoperator^{\policy}$ is defined by $(\bellmanoperator^{\policy}V)(\state) \define \expecun{}^{\policy,\prob} \bigg[ \cost(\state,\action) + \discount V(\state') \bigg ], $

where the action $a$ is chosen using $\policy$, i.e., $a\sim \pi(s)$ and the next state $\state'$ is drawn from $\prob(\cdot|\state)$.

\subsection{Value function approximation}
Most practical applications have high-dimensional state-spaces making exact computation of the value function infeasible. One solution to overcome this problem is to use a parametric approximation of the value function. In this work, we consider the linear function approximation architecture~\citep{Sutton+Barto:1998}, where the value function $V^{\policy}(\state)$, for any $\state \in \statespace$, is approximated as follows:
\begin{align}
    V^{\policy}(\state) &\approx \widetilde V(\state;\params) := \feature(\state){\tr}\params.
\end{align}
In the above, $\feature(\state) \in \real^{d}$ is a fixed feature vector for state $\state$, and $\params \in \real^{d}$ is a parameter vector that is shared across states. When the state space is a finite set, say $\statespace = \{{1}, {2}, \dots, {n}\}$, the $n$-vector $\widetilde V(\params)$ with components $\widetilde V(\state;\params)$ can be expressed as follows:
\begin{align}
   \widetilde V(\cdot;\params)&=
    \underbrace{\begin{bmatrix}
    \feature_{1}({1}) & \feature_{2}({1})& \ldots& \feature_{d}({1})\\
    \vdots & \vdots & \ddots & \vdots\\
    \feature_{1}({n})& \feature_{2}({n})& \ldots &\feature_{d}({n})
    \end{bmatrix}}_{\Feature}
    \underbrace{\begin{bmatrix}
    \params_{1}\\
    \vdots\\
    \params_{d}
    \end{bmatrix}}_{\params},
\end{align}
where $\Feature \in \real^{n \times d}$, and $\params \in \real^{d}$. 

The objective is to learn the best parameter for approximating $V^{\policy}$ within the following linear space:
\begin{align}
    \mathcal{B} := \{\Feature\params \mid \params \in \real^{d}\}.
\end{align} 
Naturally, with a linear function approximation, it is not possible to find the fixed point $V^{\policy}= \bellmanoperator^{\policy}V^{\policy}$. Instead, one can approximate $V^{\policy}$ within $\mathcal{B}$ by solving a projected system of equations. The system of equations, which is also referred to as the projected Bellman equation, is given by
\begin{align}
    \Feature\params^{\star} &= \Pi\bellmanoperator^{\policy}(\Feature\params^{\star}),\label{eq:projected_bellman}
\end{align}
where $\Pi$ is the orthogonal projection operator onto the set $\mathcal{B}$ using a weighted $\ell_2$-norm.
More precisely, let $D = \text{diag}(\stationarydist({1}), \ldots, \stationarydist({n})) \in \real^{n \times n}$ denote a diagonal matrix, whose elements are given by the stationary distribution $\stationarydist$ of the Markov chain underlying the policy $\policy$. We assume that the stationary distribution exists (see \Cref{asm:stationary}).
Let $\left\| V \right\|_D = \sqrt{V\tr D V}$ denote the weighted norm of a $n$-vector $V$, and assume that the matrix $\Feature$ has full column rank. Then, the operator $\Pi$ projects orthogonally onto $\mathcal{B}$ using the $\left\|\cdot\right\|_D$ norm, and  it can be shown that
$\Pi = \Feature (\Feature\tr D \Feature )^{-1}\Feature\tr D$.

    Next, the projected TD fixed point $\theta^*$ for \eqref{eq:projected_bellman} is given by:
    \begin{align}
        A\params^{\star} &= b,
    \textrm{ where }
        A \define \Feature{\tr}D(\identity - \discount \prob)\Feature,\quad
        b \define \Feature{\tr}D\R,\label{eq:theta-star}
    \end{align}
and $\R = \sum_{\action \in \actionspace}\policy(\state,\action)\cost(\state,\action)$.

\subsection{Temporal Difference (TD) Learning}
Temporal difference (TD)~\citep{Sutton+Barto:1998} algorithms are a class of stochastic approximation methods used for solving the projected linear system given in \eqref{eq:projected_bellman}. These algorithms start with a initial guess for the $\params_{0}$, and at every time-step $\timestep$ and update them using samples from the Markov chain induced by a policy $\policy$. 

The update rule is given as follows:
     \begin{align}
        \params_{\timestep} &= \params_{\timestep-1} + \lr f_{\timestep}(\params_{\timestep-1}),\textrm{ where} \nonumber\\
        f_{\timestep}(\params) &\define (\cost_{\timestep} +  \discount\params^{\top}\feature(\state'_{\timestep}) - \params^{\top}\feature(\state_{\timestep}))\feature(\state_\timestep).\label{eq:td-update} 
    \end{align}
    In the above, $\lr$ is the step-size parameter.

An alternate version of the algorithm (which we consider for deriving the high probability bounds) uses the projection $\Gamma$ as follows:
    \begin{align}
        \params_{\timestep} &= \Gamma(\params_{\timestep-1} + \lr f_{\timestep}(\params_{\timestep-1})). \label{eq:projected_td_update}
    \end{align}
In \eqref{eq:projected_td_update}, operator $\Gamma$ projects the iterate $\params_\timestep$ onto the nearest point in a closed ball $\mathcal{C} \in \real^d$ with a radius $H$, which is large enough to include $\params^\star$. 

An interesting result by~\citep{bvrtd} tells us that for any $\params \in \real^{d}$, the function \\$\ f(\params) \define (\cost(\state,\action) + \discount\params{\tr}\feature(\state')
    - \params{\tr}\feature(\state))\feature(\state)$ has a well defined steady-state expectation given by
\begin{align}
    \expecun{}^{\stationarydist,\prob}[f(\theta)] &= \sum_{\state,\state' \in \statespace, \action \in \actionspace}\stationarydist(\state) \pi(\state,\action)\bigg((\cost(\state,\action) + \prob(\state'|\state,\action)\discount\params{\tr}\feature(\state')
    - \params{\tr}\feature(\state))\feature(\state)\bigg).\label{eq:td_grad-a}
\end{align}

We can rearrange \eqref{eq:td_grad-a} as $ \sum_{\state,\state' \in \statespace, \action \in \actionspace}\prob(\state'|\state,\action)(\cost(\state,\state')+\discount\params{\tr}\feature(\state'_{\timestep})) = (\bellmanoperator^{\policy}\Feature\params)(s)$, and use \citep[Lemma 8]{bvrtd} to get the following:
\begin{align}
    \expecun{}^{\stationarydist,\prob}[f(\theta)]&= \Feature{\tr}D(\bellmanoperator^{\policy}\left(\Feature\params\right)-\Feature\params)\label{eq:matrix-td}\\
    &= -{A}\params + {b},\label{eq:td_grad}
\end{align}
where $A$ and $b$ are as defined in \eqref{eq:theta-star}.
We can then characterise the mean behaviour of TD algorithm using the following update rule:
    \begin{align}
       \params_{\timestep} &= \params_{\timestep-1} + \lr \bigg(\Feature{\tr}D(\bellmanoperator^{\policy}(\Feature\params_{\timestep-1})-\Feature\params_{\timestep-1})\bigg)\nonumber\\
      &= \params_{\timestep-1} + \lr \expecun{}^{\stationarydist,\prob}[f(\theta_{\timestep-1})]. \label{eq:expected-td}
    \end{align}

The characterisation of TD's behaviour in \eqref{eq:expected-td} is of particular importance as it forms the basis of our analysis.

\section{Tail-averaged TD}
\label{sec:tailavgtd}
\subsection{Basic algorithm}    
    Tail averaging or suffix averaging refers to returning the average of the final few iterates of the optimisation process, to improve its variance properties. Specifically, for any $\timestep$, the tail-averaged iterate $\params_{k+1,N}$ is the average of $\{\params_{k+1}, \ldots, \params_{t}\}$, computed as follows:
    \begin{align}
        \params_{k+1,N} &= \frac{1}{N}\sum_{i = k+1}^{k+N}\params_{i},\label{eq:tail-av-bias-var}
    \end{align}
    where $N = \timestep - k$.

Note that \citet{polyak1992acceleration}, showed that averaging all the iterates produces best asymptotic convergence rate, but from a non-asymptotic analysis viewpoint, it is usually observed that the initial error (the rate at which the initial point is forgotten) is forgotten slower with iterate averaging as compared to the non-averaged case, see \citep{fathi2013transport}. Tail averaging retains the advantages of iterate averaging, while ensuring that the initial error is forgotten exponentially fast -- a conclusion that can be inferred from the finite time bounds that we derive for the TD algorithm.

    Algorithm \ref{alg:ciac-a} presents the pseudocode of the tail-averaged TD algorithm.
    
\begin{algorithm}
    \SetKwData{Left}{left}\SetKwData{This}{this}\SetKwData{Up}{up}
    \SetKwFunction{Union}{Union}\SetKwFunction{FindCompress}{FindCompress}
    \SetKwInOut{Input}{Input}\SetKwInOut{Output}{Output}
    \SetAlgoLined
    \Input{Initial parameter $\params_{0}$,
    step-size $\lr$, initial state distribution $\zeta_{0}$, tail-average index $k$.}
    Sample an initial state $\state_{0} \sim \zeta_{0}$ \; 
    \For{ $\timestep = 0,1,\ldots$}{
        Choose an action $\action_{\timestep} \sim \policy(\state_{\timestep})$\;
        Observe $\cost_{\timestep}$, and next state $\state'_{\timestep}$\;
        Update parameters: $\params_{\timestep} = \params_{\timestep-1} + \lr f(\params_{\timestep-1})$\;
        Average the final $N$ iterates: $\params_{k+1,N} = \frac{1}{N}\sum\limits_{i = k+1}^{k+N}\params_{i}$, where $N=\timestep-k$.
     }
   \caption{Tail-averaged TD(0)}\label{alg:ciac-a}
\end{algorithm}

\subsection{Finite time bounds}
Before presenting our results, we list the assumptions under which we conduct our analysis.

    \begin{assumption}
        \label{asm:stationary}
        The Markov chain underlying the policy $\policy$ is irreducible. 
    \end{assumption}

    \begin{assumption}\label{asm:iidNoise}
        The samples $\{\state_{\timestep}, \cost_{\timestep}, \state'_{\timestep}\}_{\timestep \in \mathbb{N}}$ are independently and identically drawn from: $\stationarydist(s)\prob(s'|s)$ where, at time $\timestep$ the state $\state_{\timestep}$, where $\stationarydist$ stationary distribution underlying policy $\policy$, and $\prob(\cdot|\state_\timestep)$ is the transition probability matrix of the MDP.
    \end{assumption}
    
    \begin{assumption}\label{asm:bddFeatures}
    For all $\state \in \statespace$,  $\norm{\feature(\state)}_{2} \leq \Feature_{\mathsf{max}} < \infty$.
    \end{assumption}
    \begin{assumption}\label{asm:bddRewards}
    For all $\state \in \statespace$, and $\action\in \actionspace,$   $\left|\cost(\state,\action)\right| \leq \Cost_{\mathsf{max}}<\infty$.
    \end{assumption}
    \begin{assumption}\label{asm:phiFullRank}
    The matrix $\Phi$ has full column rank.
    \end{assumption}
    \begin{assumption}\label{asm:projection}
    The set $\mathcal{C} \define \{\params \in \real^d|\norm{\params}_2 \leq H \}$ used for projection through $\Gamma$ satisfies $H > \frac{\norm{b}_2}{\mu}$.
\end{assumption}
 We now discuss the assumptions listed above.
   \Cref{asm:stationary} ensures the existence of the stationary distribution for the Markov chain underlying policy $\policy$.
   We study the non-asymptotic behaviour of the tail-averaged TD algorithm under the i.i.d observation model as specified in \Cref{asm:iidNoise}, and later show that our results can be extended to handle Markov sampling.
    Next, Assumptions \ref{asm:bddFeatures} and \ref{asm:bddRewards} are boundedness requirements on the underlying features and rewards, and are common in the finite time analysis of TD algorithm, see \citep{Bhandari0S18,lstd-prashanth}.
    Assumption \ref{asm:phiFullRank} requires the columns of the feature matrix $\Feature$ to be linearly independent, in turn ensuring the uniqueness of the TD solution $\params^*$. Moreover, this assumption ensures that the minimum eigenvalue, say $\mu'$ of $B=\expecun{}^{\stationarydist,\prob}[\Feature\Feature\tr]$ is strictly positive, in turn implying that the minimum eigenvalue $\mu$ of the matrix $A$ defined in \eqref{eq:theta-star} is strictly positive. \Cref{asm:projection} is required for the high-probability bounds, while the bounds in expectation do not require projection.
    
    The first result we state below is a bound in expectation on the parameter error $\norm{\params_{k+1,N} - \params^{\star}}^{2}_{2}$.

    \begin{theorem}[\textbf{Bound in expectation}] \label{thm:expectation_bound}
      Suppose  \Crefrange{asm:stationary}{asm:phiFullRank} hold.  Choose a step size $\lr$ satisfying
        \begin{align*}
        \lr \leq \lr_{\mathsf{max}} =  \frac{1-\discount}{(1+\discount)^2\Phi^{2}_{\mathsf{max}}},
        \numberthis\label{eq:gammamax}
        \end{align*}
        where $\beta$ is the discount factor and $\Phi_{\mathsf{max}}$ is a bound on the features (see Assumption \ref{asm:bddFeatures}).
        
        Then the expected error of the tail-averaged iterate $\params_{k+1,N}$ when using Algorithm \ref{alg:ciac-a} satisfies
       \begin{align*}
       \expecun{}\left[\norm{\params_{k+1,N} - \params^{\star}}^{2}_{2}\right] &\leq \frac{10  e^{(-k \lr (1-\discount)\mu')}}{\lr^2 (1-\discount)^2\mu'{}^{2} N^{2}}\expecun{}\left[\norm{\params_{0} - \params^{\star}}^{2}_2\right] + \frac{10\sigma^{2}}{(1-\discount)^2\mu'{}^2 N},\numberthis\label{eq:td-expec-bd}
    \end{align*}
       where $N = \timestep - k$, $\theta_0$ is the initial point, $\sigma^2 = (\Cost_{\mathsf{max}} + (1+\discount)\Feature^{2}_{\mathsf{max}}\norm{\params^{\star}}_{2}^{2})$,  with $\theta^\star$ denoting the TD fixed point specified in \eqref{eq:theta-star}, and $\mu'$ is the minimum eigenvalue of $B=\expecun{}^{\stationarydist,\prob}[\Feature\Feature\tr]$.
    \end{theorem}
    \begin{proof}
        See Section \ref{sec:proofsketch-expectationbound} for a sketch and \Cref{proof:thm-1} for a detailed proof.
    \end{proof}
    A few remarks are in order.
\begin{remark}
It is apparent that the bound presented above scales inversely with the square of $(1-\beta)\mu'$.
More importantly, the bound presented above is for a step-size choice that does not require  information about the eigenvalues of matrices $A$ or $B$. To the best of our knowledge, this is the first bound of $O\left(1/t\right)$ for a `universal' step-size. Previous results, such as those by \citep{Bhandari0S18,lstd-prashanth} provide a comparable bound, albeit for a diminishing step-size of the form $c/k$, where setting $c$ requires knowledge of $\mu$. On the other hand, \citep{DalalSTM18,lstd-prashanth} provide a $O\left(1/t^\alpha\right)$ bound for larger step-sizes of the form $c/t^\alpha$, where $c$ is a universal constant. 
\end{remark}
\begin{remark}
	The first term on the RHS of \eqref{eq:td-expec-bd} relates to the rate at which the initial parameter $\params_0$ is forgotten, while the second term arises from a martingale difference noise term associated with the i.i.d. sampling model.  
	Setting $k=t/2$, we observe that the first term is forgotten at an exponential rate, while the noise term is $O(1/t)$.
\end{remark}
\begin{remark}
	In \citep{lakshminarayanan18a}, the authors consider iterate averaging in a linear stochastic approximation setting. Comparing their Theorem 1 to the result we have presented above, we note that the first term on the RHS of \eqref{eq:td-expec-bd} exhibits an exponential decay, while the corresponding decay is of order $O(1/t)$ in \citep{lakshminarayanan18a}. The second term in their result as well as in \eqref{eq:td-expec-bd} is of order $O(1/t)$. While the second dominates the rate, the first term, which relates to the rate at which the initial parameter is forgotten, decays much faster with tail averaging. Intuitively, it makes sense to average after sufficient iterations have passed, instead of averaging from the beginning, and our bounds confirm this viewpoint.
\end{remark}	
\begin{remark}
	A closely related result under comparable assumptions is Theorem 2 of \citep{Bhandari0S18}. This result provides two bounds corresponding to constant and diminishing step-sizes, respectively, while assuming the knowledge of $\mu$. The bound there corresponding to the constant step-size for the last iterate of TD is the sum of an exponentially decaying `initial error' term and a constant offset with the noise variance. The second bound in the aforementioned work is $O(1/t)$ for both initial error and noise terms. 
	The bound we derived in \eqref{eq:td-expec-bd} combines the best of these two bounds through tail averaging, i.e., an exponentially decaying initial error, and a $O(1/t)$ noise term. As an aside, our bound is for the projection-free variant of TD, while the bounds in \citep{Bhandari0S18} requires projection, with an assumption similar to Assumption \ref{asm:projection}.
\end{remark}
\begin{remark}
	Another closely related result is Theorem 4.4 of \citep{lstd-prashanth}, where the authors analyse TD with linear function approximation, with input data from a batch of samples. The analysis there can be easily extended to cover our i.i.d. sampling model. As in the remark above, while the overall rate is $O(1/t)$ in their result as well as \eqref{eq:td-expec-bd}, the initial error in our bound is forgotten much faster. A similar observation also holds w.r.t. the bound in the recent work \citep{devraj}, but the authors do not state their bound explicitly. 
\end{remark}
\begin{remark}
	It is possible to extend our analysis to cover the Markov noise observation model, as specified in Section 8 of \citep{Bhandari0S18}. 
	In this model, we assume that the underlying Markov chain is uniformly ergodic, which intuitively translates to a fast mixing rate. For finite Markov chains irreducibility and aperiodicity are sufficient to establish this assumption. The fast mixing assumption allows us to translate the i.i.d. sampling resutls to Markov sampling. We provide the details of such an extension in Section \ref{sec:mixing}.
\end{remark}

Next, we turn to providing a bound that holds with high probability the parameter error $\norm{\params_{k+1,N} - \params^{\star}}^{2}_{2}$ of the projected TD algorithm. For this result, we require the TD update parameter to stay within a bounded region that houses $\params^*$, which is formalized in  \Cref{asm:projection}.
    
\begin{theorem}[\textbf{High-probability bound}]\label{thm:high-prob-bound}
    	Suppose \Crefrange{asm:stationary}{asm:projection} hold. Choose the step size such that $\lr \leq \lr_{\mathsf{max}}$, where $\lr_{\mathsf{max}}$ is defined in \eqref{eq:gammamax}. Then, for any  $\delta\in (0,1]$, we have the following bound for the projected tail-averaged iterate $\params_{k+1,N}$:
    \begin{align*}
        \prob\left(\norm{\params_{k+1,N} - \params^\star}_{2} \leq \frac{2\sigma}{(1-\discount)\mu' \sqrt{N}}\sqrt{\log\left(\frac{1}{\delta}\right)} +  \frac{4 e^{(-k \lr (1-\discount)^{2}\mu')}}{\lr (1-\discount)\mu'N}\expecun{}\left[\norm{\params_{0} - \params^\star}_{2}\right] + \frac{4\sigma}{(1-\discount)\mu' \sqrt{N}} \right) &\geq 1 - \delta,
    \end{align*}
    where $N,\sigma,\mu, \theta_0,\theta^\star$ are as specified in Theorem \ref{thm:expectation_bound}.
\end{theorem}
    \begin{proof}
        See \Cref{sec:proofsketch-highprobbound} for a sketch and \Cref{proof:hpb-1} for a detailed proof.
    \end{proof}

\begin{remark}
	High-probability bounds for TD algorithm have been derived earlier in \citep{lstd-prashanth,DalalSTM18}. In comparison to Theorem 4.2 of \citep{lstd-prashanth}, we note that our bound is an improvement since the sampling error (the first and third terms in $\K(n)$ defined above) decays at a much faster rate for tail-averaged TD. 
	Next, unlike \citep{DalalSTM18}, we note that our bound requires projection. However it does exhibit a $O(1/\timestep)$ rate. The result by \citep{DalalSTM18} (Theorem 3.6) is of the form $O(1/\timestep^\lambda)$ where $\lambda$ is related to the minimum eigenvalue $\mu$ of matrix $A$, and hence cannot be guaranteed to be of order $O(1/\timestep)$.
\end{remark}

\begin{remark}\label{rem:markov}
Consider the case when Assumption~\ref{asm:iidNoise} does not hold, but we sample $\left(s_t,r_t,s_t^{\prime}\right)$ from a trajectory corresponding the policy $\pi$.  We assume exponential ergodicity for the total variation distance as used in \citep{Bhandari0S18,srikant19a}, with mixing time $\tmix$. For this case, we consider a variant of TD which uses one sample in every $\tilde{O}\left(\tmix\right)$ consecutive samples for the update iteration.
The guarantees for the resulting TD algorithm with tail averaging with $N$ data points in the trajectory corresponds to the guarantees for $\theta_{k+1,N^{\prime}}$ in Theorem~\ref{thm:high-prob-bound} where $N^{\prime} = \tilde{O}\left(\tfrac{N}{\tmix}\right)$, and $(1-\delta)$ is replaced by $(1-2\delta)$. This gives an error of the order $\tilde{O}\left(\sqrt{\frac{\tmix}{N}}\right)$, which is similar to the bounds in \citep[Theorem 3]{Bhandari0S18}. These follow from standard mixing arguments and we refer to Section~\ref{sec:mixing} for further details.
As an aside, we remark that one cannot get a better bound from an information-theoretic viewpoint without further assumptions on the nature of the linear approximation (cf. Theorem 2 in \citep{nagaraj2020least}).
\end{remark}      

\section{Regularized TD Learning}
\label{sec:regtd}
In this section, we analyse the regularised TD algorithm. From the results in \Cref{thm:expectation_bound,thm:high-prob-bound} one can observe that although tail-averaged TD achieves a $O\left(\frac{1}{t}\right)$ rate of convergence, the bounds depend inversely on $(1-\beta)\mu'$, where $\mu'$ is the  minimum eigenvalue of $B=\expecun{}^{\stationarydist,\prob}[\Feature\Feature\tr]$. 
In the following results we will show that the non-asymptotic bounds for regularised TD scale inversely with $\mu$ (minimum eigenvalue of matrix $A$). Such a dependence may be preferable over vanilla TD, as there are problem instances where $(1-\beta)\mu' \ll \mu$. To make this intuition more concrete, consider the following problem instance.

\begin{example}
\begin{figure}[h]
    \centering
\begin{tikzpicture}[->, >=stealth', auto, semithick, node distance=3cm]
	\tikzstyle{every state}=[fill=white,draw=black,thick,text=black,scale=1]
	\node[state]    (A)                     {$0$};
	\node[state]    (B)[right of=A]   {$1$};
	\path
	(A) edge[loop left]			node{$1-p$}	(A)
	edge[bend left,above]		node{$p$}	(B)
	(B) edge[bend left,below]	node{$p$}	(A)
	edge[loop right]		node{$1-p$}	(B);
	\end{tikzpicture}
    \caption{A two state Markov chain}
    \label{fig:mc}
\end{figure}
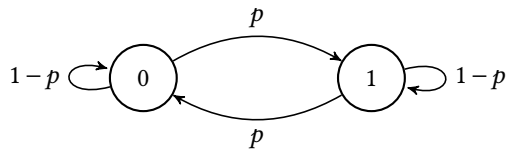
Consider a two state MDP with the transition dynamics as depicted in Figure \ref{fig:mc}, for a given policy, say $\pi$. The one-dimensional state features are given as follows: $\phi(1) = 1$, and $\phi(2) = \frac{1}{2}$. For the case of $p=\frac{1}{2}$, we have
\[A=(1/2)\left(\phi(1)^2 + \phi(2)^2\right) - \beta/4 \left(\phi(1)^2 +\phi(2)^2  + \phi(1)\phi(2) + \phi(2)\phi(1)\right)=\frac{5}{8}-\frac{9\beta}{16}.\] Further, $B=\frac{5}{8}$. Thus, for any $\beta\in [0,1]$, 
$(1-\beta)B \le A$. Further, as $\beta$ approaches $1$, $(1-\beta B)\rightarrow 0$, while $A\rightarrow \frac{1}{16}$. Since convergence rate of tail-averaged TD depends inversely on $(1-\beta)\mu'$ (see Theorems \ref{thm:expectation_bound} and \ref{thm:high-prob-bound}), there is a concrete case for an algorithm whose convergence rate depends on $\mu$ instead of $(1-\beta)\mu'$. The regularised TD variant that we present next achieves this objective.
\end{example}

\subsection{Basic algorithm}
Instead of the TD solution \eqref{eq:td-solution}, we solve the following regularised problem for a given regularisation parameter $\lambda>0$:
\begin{align}
    \params^{\star}_{\mathsf{reg}} &= (A + \lambda \identity)^{-1}b, \label{eq:sol-reg-td}
\end{align}

The update iteration of the TD analogue for the regularised case is as follows:
\begin{align}
    \hat{\params}_\timestep &= (\identity- \lr \lambda)\hat{\params}_{\timestep-1} + \lr (\cost_{\timestep} +  \discount\hat{\params}^{\top}_{\timestep-1}\feature(\state'_{\timestep}) - \hat{\params}^{\top}_{\timestep-1}\feature(\state_{\timestep}))\feature(\state_\timestep).\label{eq:req-td-update}
\end{align}

Similarly, the projected regularised TD update (which we consider for deriving the high probability bounds) uses the projection $\Gamma$ as follows:
    \begin{align}
        \hat{\params}_\timestep &= \Gamma((\identity - \lr \lambda)\hat{\params}_{\timestep-1} + \lr (\cost_{\timestep} +  \discount\hat{\params}^{\top}_{\timestep-1}\feature(\state'_{\timestep}) - \hat{\params}^{\top}_{\timestep-1}\feature(\state_{\timestep}))\feature(\state_\timestep)). \label{eq:regprojected_td_update}
    \end{align}
In \eqref{eq:regprojected_td_update}, operator $\Gamma$ projects the iterate $\params_\timestep$ onto the nearest point in a closed ball $\mathcal{C} \in \real^d$ with a radius $H$ which is large enough to include $\params^{\star}_{\mathsf{reg}}$. 

Using arguments similar to vanilla TD, it is easy to see that the iterate $\hat\theta_t$ converges \eqref{eq:sol-reg-td} under   \Crefrange{asm:stationary}{asm:phiFullRank}, and a standard stochastic approximation condition on the step-size. 

The overall flow of the regularised TD algorithm would be similar to Algorithm \ref{alg:ciac-a}, except that the iterate is updated according to \eqref{eq:req-td-update}, and an additional regularisation parameter is involved.


\subsection{Finite time bounds}
Using a technique similar to that used in establishing the bound for tail-averaged TD in Theorem \ref{thm:expectation_bound}, we arrive at the following bound in expectation for regularised TD. 
    \begin{theorem}[\textbf{Bound in expectation}] \label{thm:regtd-expectation_bound}
          Suppose  \Crefrange{asm:stationary}{asm:bddRewards} hold.  
            Choose a step size $\lr$ satisfying 
            \begin{align*}
               \lr \leq \lr_{\text{max}} = \frac{\lambda}{\lambda^2 + 2\lambda (1+\discount)\Phi^{2}_{\mathsf{max}} + (1+ \discount)^2 \Phi^4_{\mathsf{max}}}.\numberthis\label{eq:reg-td-lr}
            \end{align*}
            Then the expected error of the tail-averaged regularised TD iterate $\hat{\params}_{k+1,N}$ satisfies
        \begin{align}
            \expecun{}\left[\norm{\hat{\params}_{k+1,N} - \params^{\star}_{\mathsf{reg}}}^{2}_{2}\right] &\leq \frac{  10e^{(-k \lr (\mu+\lambda))}}{\lr^2  (\mu+\lambda)^{2} N^{2}}\expecun{}\left[\norm{\hat{\params}_{0} - \params^{\star}_{\mathsf{reg}}}^{2}_2\right] + \frac{10\sigma^{2}}{(\mu+\lambda)^{2} N},\label{eq:regtd-expec-bd}
        \end{align}
           where $N = \timestep - k$, and $\sigma^2 = (\Cost_{\mathsf{max}} + (1+\discount)\Feature^{2}_{\mathsf{max}}\norm{\params^{\star}_{\mathsf{reg}}}_{2}^{2})$.
    \end{theorem}
    \begin{proof}
         See Section \ref{sec:proofsketch-regtdexpec} for a sketch and \Cref{sec:expectation-bound} for a detailed proof.
    \end{proof}

While the result above bounded the distance to the regularised TD solution, the next result shows that the distance between regularised TD iterate and vanilla projected TD fixed point is of $O\left(\lambda\right)$.    
\begin{corollary}
\label{cor:reg-td}
Under conditions of Theorem \ref{thm:regtd-expectation_bound}, we have
        \begin{align*}
             \expecun{}\left[\norm{\hat\params_{k+1,N} - \params^{\star}}_{2}^{2}\right] &\leq  \frac{  20e^{(-k \lr (\mu+\lambda))}}{\lr^2 (\mu+\lambda)^2 N^{2}}\expecun{}\left[\norm{\hat{\params}_{0} - \params^{\star}_{\mathsf{reg}}}^{2}_2\right] + \frac{20\sigma^{2}}{(\mu+\lambda)^2 N} + \frac{2 \lambda^2 \Feature^{2}_\mathsf{max}R_\mathsf{max}^{2}}{\sigma_{\mathsf{min}}(A)^2(\mu+\lambda)^2},\numberthis\label{eq:corollary}
        \end{align*}
    where $\sigma_{\mathsf{min}}(A)$ is $A$'s minimum singular value
\end{corollary}
\begin{proof}
    See \Cref{proof:cor1}
\end{proof}
With a suitable choice of $\lambda$, the next result shows that regularised TD obtains a $O\left(1/t\right)$ rate (e.g., with $k=t/2$), and the bound scales inversely with the eigenvalue $\mu$ of the matrix $A$. From the discussion earlier, recall that there are problem instances where $\mu \gg (1-\beta)\mu'$, and the bound for tail-averaged TD sans regularisation depended inversely on $(1-\beta)\mu'$.
\begin{corollary}
\label{cor:reg-td-lambda}
Under conditions of Theorem \ref{thm:regtd-expectation_bound}, and with $\lambda = \frac{1}{\sqrt{N}}$, we obtain
        \begin{align*}
             \expecun{}\left[\norm{\hat\params_{k+1,N} - \params^{\star}}_{2}^{2}\right] &\leq \frac{  20(1+(1+\discount)\Feature_\mathsf{max}^2 \sqrt{N})^{4}e^{\frac{(-k\mu)}{(1+\discount)^2\Feature_\mathsf{max}^4\sqrt{N}}}}{\mu^2 N^{3}}\expecun{}\left[\norm{\hat{\params}_{0} - \params^{\star}_{\mathsf{reg}}}^{2}_2\right] + \frac{20\sigma^{2}}{\mu^2 N} + \frac{2 \Feature^{2}_\mathsf{max}R_\mathsf{max}^{2}}{\mu^2 N}\numberthis\label{eq:corollary-2}
        \end{align*}
\end{corollary}


A few remarks are in order.
\begin{remark}
The choice of step-size in the bound of Theorem \ref{thm:regtd-expectation_bound} is universal, i.e., does not require the knowledge of $\mu$, and the rate of convergence is $O(1/t)$, if we set $k=t/2$, or any constant multiple of $t$. However, the regularised TD iterate converges to \eqref{eq:sol-reg-td}, which is different from the vanilla TD fixed point. But, the distance is between the regularised and vanilla TD solutions is $O(\lambda)$, implying that for small value of $\lambda$, the regularised TD solution is a good proxy, which in turn implies that the regularised TD iterate can be used in place of vanilla TD iterate, to obtain a good approximation to the TD fixed point. Corollary \ref{cor:reg-td} makes this intuition precise.     
\end{remark}

\begin{remark}
The initial and sampling errors in \eqref{eq:regtd-expec-bd} are as in the tail-averaged TD (see Theorem \ref{thm:expectation_bound}), i.e., initial error is forgotten at an exponential rate, while the sampling error is $O(1/t)$. 
\end{remark}    

\begin{remark}
In \citep{lstd-prashanth}, the authors analyse the iterate-average variant of TD, and derive a $O(1/t^\alpha)$ bound for a step-size $\Theta(1/k^\alpha)$, where $1/2< \alpha<1$. Further, their step-size choice is universal as is the case of tail-averaged TD in comparison to \citep{lstd-prashanth}. Our bound for regularised TD exhibits a better rate, though the bound measures the distance to the regularised TD solution. 
\end{remark}

Next, we present a high-probability bound for regularised TD in the spirit of Theorem \ref{thm:high-prob-bound}.
\begin{theorem}[\textbf{High-probability bound}]\label{thm:regtd-high-prob}
    		Suppose \Crefrange{asm:stationary}{asm:bddRewards}, and \ref{asm:projection} hold. Choose the step size such that $\lr \leq \lr_{\mathsf{max}}$, where $\lr_{\mathsf{max}}$ is defined in \eqref{eq:reg-td-lr}. Then, for any  $\delta\in (0,1]$, we have the following bound for the projected tail-averaged regularised TD iterate $\hat{\params}_{k+1,N}$:
    \begin{align*}
        \prob\left(\norm{\hat{\params}_{k+1,N} - \params^{\star}_{\mathsf{reg}}}_{2} 
        \leq 
        \frac{2\sigma}{(\mu+\lambda)\sqrt{N}}\sqrt{\log\left(\frac{1}{\delta}\right)} +  \frac{4e^{(-k \lr (\mu + \lambda))}}{\lr(\mu+\lambda) N}\expecun{}\left[\norm{\hat{\params}_{0} - \params^{\star}_{\mathsf{reg}}}_{2}\right] +  \frac{4\sigma}{(\mu + \lambda)\sqrt{N}}\right) &\geq 1 - \delta,
    \end{align*}
    where $N,\sigma,\mu, \hat{\params}_0,\params^{\star}_{\mathsf{reg}}$ are as specified in Theorem \ref{thm:regtd-expectation_bound}.
\end{theorem}
    \begin{proof}
        The proof follows the same template as Theorem \ref{thm:high-prob-bound}, and is given in \Cref{proof:thm:regtd-high-prob}.
    \end{proof}

As discussed in Remark \ref{rem:markov} and Section \ref{sec:mixing} for tail-averaged TD sans regularisation, it is straightforward to extend the results in Theorems \ref{thm:regtd-expectation_bound} and \ref{thm:regtd-high-prob} to cover the case of Markov sampling. 

\section{Proof Ideas}
\label{sec:proofsketch}
\subsection{Proof of Theorem \ref{thm:expectation_bound} \textit{(Sketch)}}
\label{sec:proofsketch-expectationbound}
\begin{proof}
    We present here the framework for obtaining the results obtained in the paper; the framework has been introduced in the work of~\citep{lstd-prashanth,Bhandari0S18,DalalSTM18}. Towards that end, we begin by introducing some notation. First, we define the centered error $z_\timestep\define \params_{\timestep} - \params^{\star}$.
    Using the TD update \eqref{eq:td-update}, the centered error can be seen to satisfy the following recursive relation:
    \begin{align}
        z_{\timestep} &= (\identity - \lr a_\timestep) z_{\timestep-1} + \lr f_{\timestep}(\params^{\star}),
    \end{align}
    where $f(\cdot)$ defined as in \eqref{eq:td-update}, and $a_\timestep \define \feature(\state_{\timestep})\feature(\state_{\timestep}){\tr} - \discount\feature(\state_{\timestep})\feature(\state'_{\timestep}){\tr}$. 
    
     The centered error is decomposed into a bias and variance term as follows:
     \begin{align*}
         \expecun{}\left[\norm{z_\timestep}_{2}^{2}\right] &= 2\expecun{}\left[\norm{C^{\timestep:1} z_0]}_{2}^{2}\right] + 2\lr^2 \sum_{k=0}^{\timestep} \expecun{}\left[\norm{ \sum_{k=0}^{\timestep} C^{t:k+1}f_{k}(\params^\star)}_{2}^{2}\right]\\
         &= 2 z_{\timestep}^{\text{bias}} + 2 \lr^2 z_{\timestep}^{\text{variance}},
     \end{align*}
     where 
    \begin{equation}
        C^{i:j}=
        \begin{cases}
          (\identity - \lr a_i) (\identity - \lr a_{i-1})\ldots (\identity - \lr a_j) , & \text{if}\  i >= j \\
          \identity, & \text{otherwise}
        \end{cases}
    \end{equation}

     The bias term is then bounded as $z_{\timestep}^{\text{bias}} \leq \exp(-\lr(1-\discount)\mu'\timestep)\expecun{}[\norm{z_{0}}^{2}_{2}]$, while the variance term is bounded as $z_{\timestep}^{\text{variance}} \leq \frac{\sigma^{2}}{(1-\discount)\mu'}$.
     
     The centered error corresponding to the tail-averaged iterate $\params_{k+1,N}$ is given by $z_{k+1,N} = \frac{1}{N}\sum_{i = k+1}^{k+N}z_{i}$. The analysis proceeds by bounding the expectation of the norm $\expecun{}[\norm{z_{k+1,N}}_{2}^{2}]$ using the following decomposition:
     \begin{align*}
         \expecun{}\left[\norm{z_{k+1,N}}_{2}^{2}\right]
                                  &\leq \frac{1}{N^{2}}\bigg(\sum_{i= k+1}^{k+N}\expecun{}\left[\norm{z_i}_{2}^{2}\right] + 2 \sum_{i=k+1}^{k+N-1}\sum_{j=i+1}^{k+N} \expecun{}\left[z_{i}^{\top}z_{j}\right] \bigg).
    \end{align*}
    Using the definitions of $z_{\timestep}^{\text{bias}}$ and $z_{\timestep}^{\text{variance}}$, we simplify the RHS above as follows:
     \begin{align}
             \expecun{}\left[\norm{z_{k+1,N}}_{2}^{2}\right] &\leq \underbrace{\frac{2}{N^{2}}\bigg(1+\frac{4}{\lr(1-\discount) \mu'}\bigg) \sum_{i = k+1}^{k+N}   z_{i}^{\text{bias}}}_{z_{k+1, N}^{\text{bias}}} + \underbrace{\frac{2}{N^{2}}\bigg(1+\frac{4}{\lr(1-\discount) \mu'}\bigg)\lr^{2}  \sum_{i = k+1}^{k+N}   z_{i}^{\text{variance}}}_{z_{k+1,N}^{\text{variance}}}\label{eq:tail-av-bias-var-dec},
    \end{align}
    where, $z_{i}^{\text{bias}} = \expecun{}\left[\norm{C^{\timestep:1} z_0]}_{2}^{2}\right] $ and $z_{i}^{\text{variance}} = \sum_{k=0}^{\timestep} \expecun{}\left[\norm{\sum_{k=0}^{\timestep} C^{t:k+1}f_{k}(\params^\star)}_{2}^{2}\right]$, and $f(\cdot)$ defined as in \eqref{eq:td-update}.
    
    The main result follows by substituting the bounds on $z_{i}^{\text{bias}}$ and $z_{i}^{\text{variance}}$ followed by some algebraic manipulations.
\end{proof}

\subsection{Proof of Theorem \ref{thm:high-prob-bound} \textit{(Sketch)}}\label{sec:proofsketch-highprobbound}
    \begin{proof}
    To obtain the high-probability bound we use the proof technique by \citet{lstd-prashanth}, where we consider separately the deviation of the centered error from its mean,~\ie $\norm{z_{k+1,N}}_{2}^{2} - \expecun{}\left[\norm{z_{k+1,N}}_{2}^{2}\right]$. We decompose this quantity as a sum of martingale differences, establish a Lipschitz property followed by a sub-Gaussian concentration bound to infer 
    \begin{align}
            \prob\left(\norm{z_{k+1,N}}_{2} - \expecun{}\left[\norm{z_{k+1,N}}_{2}\right] > \epsilon\right) \leq \exp\left(-\frac{\epsilon^{2}}{ (\Cost_{\text{max}} + (1+ \discount) H \Feature^{2}_{\text{max}})^{2} \sum_{i = k+1}^{k+N} L^{2}_{i}} \right),\label{eq:hpb-1}
    \end{align}
    where $L_{i} \define \frac{\lr}{N}\sum_{j=i+1}^{i+N}\left(1 - \frac{\lr(1-\discount)\mu'}{2}\right)^{j-i+1}$
    Next, under the choice of step-size $\lr$ specified in the theorem statement, we establish that
    \begin{align*}
            \sum_{i=k+1}^{k+N} L_{i}^{2} &\leq \frac{4}{N(1-\discount)\mu'^2}.
    \end{align*}
    The main claim follows by (i) substituting the bound obtained above in \eqref{eq:hpb-1}; (ii) using the bound on $\expecun{}\left[\norm{z_{k+1,N}}_{2}\right]$ specified in \Cref{thm:expectation_bound}; and (iii) converting the tail bound resulting from (i) and (ii) into a high-probability bound.
    
    
    
    
    
    The detailed proof is given in \Cref{sec:high-prob}.
    \end{proof}

\subsection{Proof of Theorem \ref{thm:regtd-expectation_bound} (\textit{Sketch})}\label{sec:proofsketch-regtdexpec}
The proof of \Cref{thm:regtd-expectation_bound} follows the same template as \Cref{thm:expectation_bound}. The following lemma captures the interplay of the step size and regularization parameters and its subsequent effect on the constants and decay rates in \Cref{thm:regtd-expectation_bound}'s result.
\begin{lemma}\label{lem:regtd-featurebound-main}
    With $\lr \leq \lr_{\text{max}}$ as given in \eqref{eq:reg-td-lr}, 
    the following bound holds
    \begin{equation*}
         \bigg\Vert\bigg(\identity - \lr(A + \lambda\identity)\bigg)^{\top} \bigg(\identity - \lr(A + \lambda\identity)\bigg)\bigg\Vert_2 \leq 1 - \lr(\mu+\lambda),\\
         \quad \text{and}\quad
         \Vert(\identity - \lr A)\Vert_2 \leq 1 - \frac{\lr(\mu+\lambda)}{2}.
    \end{equation*}
\end{lemma}
\section{Convergence Analysis}\label{sec:convergence-analysis}




  \subsection{Preliminaries }\label{sec:convergence-proofs}
    Let $\mathcal{F}_{\timestep}$ denote the sigma-field generated by $\{\params_0 \ldots \params_\timestep\}, \ \timestep \geq 0 $, and let
    \begin{align*}
        f_{\timestep}(\params) \define (\cost_{\timestep} +  \discount\params^{\top}\feature(\state'_{\timestep}) - \params^{\top}\feature(\state_{\timestep}))\feature(\state_\timestep). \numberthis\label{eq:f}
    \end{align*}
    Recall that $n$ denotes the number of states in the underlying MDP and the feature matrix $(|\statespace| \times d)$-matrix, where $\Feature \define (\feature_{1}(\state_1),\ldots, \feature_{d}(\state_{n}))$.
    
    \noindent According to the characterisation of TD's steady-state behaviour in \eqref{eq:expected-td}, it's final solution can be written as follows:
    \begin{align*}
        \params^\star = A^{-1}b, \numberthis\label{eq:td-solution}
    \end{align*}
    Using \eqref{eq:expected-td}, we rewrite the TD update in \cref{eq:td-update} as follows:
    \begin{align*}
        \params_\timestep &= \params_{\timestep-1} + \lr f_{\timestep}(\params_{\timestep-1}) \\
        &= \params_{\timestep-1} +\lr\bigg(-A\params_{\timestep-1} + b + \Delta M_{\timestep}\bigg), \numberthis\label{eq:modified-td-update}
    \end{align*}
    where $\Delta M_\timestep = f_{\timestep}(\params_{\timestep-1}) - \expecun{}^{\stationarydist,\prob}[f_{\timestep}(\params_{\timestep-1})|\mathcal{F}_{\timestep-1}]$ is a martingale difference sequence with $f$ defined as in \eqref{eq:f}.
    
    \vspace{0.5cm}
        \noindent We shall first establish a few useful results in \Cref{lem:A-upper,lem:A-lower,lem:A-upper-2,lem:high-prob-helper-1}.
    \vspace{0.2cm}  
    
        \begin{lemma}\label{lem:psd-bound}
            For any $a,b \in \real^d$, we have
            \begin{align*}
             -\frac{\params\tr(aa\tr + bb\tr)\params }{2} \leq   \params\tr (ab\tr) \params \leq \frac{\params\tr(aa\tr + bb\tr)\params }{2}.
            \end{align*}
        \end{lemma}
        \begin{proof}
             \begin{align*}
                 \params\tr ab\tr \params &\stackrel{(a)}{\leq} \bigg( \left(\params\tr a a\tr \params\right)\left(\params \tr b b\tr \params \right)\bigg)^{1/2}\\
                 &\stackrel{(b)}{\leq} \frac{\left(\params\tr a a\tr \params\right) + \left(\params \tr b b\tr \params \right) }{2}\\
                 &= \frac{\params\tr(aa\tr + bb\tr)\params}{2},
             \end{align*}
        where $(a)$ follows from Cauchy-Schwarz inequality and $(b)$ follows from AM-GM inequality.
        \end{proof}

        \begin{lemma}\label{lem:A-upper}
        The matrix $A$ defined in \eqref{eq:td-solution} satisfies
        \begin{align*}
           \Vert A \Vert_2 &\leq (1+\discount)\Feature_{\mathsf{max}}^2, 
        \end{align*}
        where $\Feature_{\mathsf{max}}$ is specified in \Cref{asm:bddFeatures}, and $\beta$ is the discount factor.
        \end{lemma}
        \begin{proof}
            \begin{align*}
                \Vert A \Vert_2 &= \Vert \expecun{}[\feature(\state_\timestep)\feature(\state_\timestep)\tr - \discount \feature(\state_\timestep)\feature(\state'_\timestep)\tr] \Vert_2\\
                &\stackrel{(a)}{\leq} \expecun{}[\Vert\feature(\state_\timestep)\feature(\state_\timestep)\tr\Vert_2] + \discount \expecun{}[\Vert \feature(\state_\timestep)\feature(\state'_\timestep)\tr\Vert_2]\\
                &\stackrel{(b)}{\leq} (1+\discount)\Feature_{\mathsf{max}}^2,\\
            \end{align*}
            where $(a)$ is due to Jensen's inequality, and $(b)$ uses \Cref{asm:bddFeatures}.
         \end{proof}

        \begin{lemma}\label{lem:A-lower}
        Recall that $B \define \expecun{}^{\stationarydist}[\feature(\state_\timestep)\feature(\state_\timestep)\tr]$, For any $\params \in \real^{d}$ the following inequality holds
            \begin{align*}
              2(1-\discount)\params\tr B \params \leq  \params\tr(A + A\tr)\params \leq  2 (1+\discount) \params\tr B \params.
            \end{align*}
        \end{lemma}
        \begin{proof}
        \noindent We will first provide a proof for the lower bound
             \begin{align*}
                 \params\tr(A + A\tr)\params &= \params\tr\left(\expecun{}[\feature(\state_\timestep)\feature(\state_\timestep)\tr - \discount \feature(\state_\timestep)\feature(\state'_\timestep)\tr] + \expecun{}[(\feature(\state_\timestep)\feature(\state_\timestep)\tr)\tr - \discount (\feature(\state_\timestep)\feature(\state'_\timestep)\tr)\tr]\right)\params \\
                 &= \params\tr(2B - \discount \expecun{}[\feature(\state_\timestep)\feature(\state'_\timestep)\tr + \feature(\state'_\timestep)\feature(\state_\timestep)\tr])\params\numberthis\label{eq:Alow-interim}\\
                 & \stackrel{(a)}{\geq} \params\tr(2B - \discount( \expecun{}[\feature(\state_\timestep)\feature(\state_\timestep)\tr ] + \expecun{}[\feature(\state'_\timestep)\feature(\state'_\timestep)\tr ])\params \\
                 &\stackrel{(b)}{\geq} 2(1-\discount)\params\tr B \params,
             \end{align*}
            where $(a)$ is uses \Cref{lem:psd-bound}, and $(b)$ follows from the fact that $\expecun{}^{\stationarydist}[\feature(\state_\timestep)\feature(\state_\timestep)\tr]  = \expecun{}^{\stationarydist}[\feature(\state'_\timestep)\feature(\state'_\timestep)\tr]$ which holds because $\feature(s_\timestep)$ and $\feature(s^{'}_{\timestep})$ are both sampled from the stationary distribution $\stationarydist$.

           \noindent  For the upper bound, we use \Cref{lem:psd-bound} \eqref{eq:Alow-interim} to obtain
            \begin{align*}
                \params\tr(A + A\tr)\params &\leq \params\tr(2B + \discount \expecun{}[\feature(\state_\timestep)\feature(\state_\timestep)\tr + \feature(\state'_\timestep)\feature(\state'_\timestep)\tr])\params\\
                &= 2(1+\discount)\params\tr B\params.
            \end{align*}
        \end{proof}
       
        \begin{lemma}\label{lem:A-upper-2}
        Let $a_{j} \define [\feature(\state_{j})\feature(\state_{j}){\tr} - \discount\feature(\state_{j})\feature(\state'_{j}){\tr}]$, and $B \define \expecun{}^{\stationarydist}[\feature(\state_\timestep)\feature(\state_\timestep)\tr]$. Then for any $\params \in \real^{d}$, the following inequality holds
            \begin{align*}
                \params\tr(\expecun{}[a_{j}\tr a_{j}])\params \leq \Feature^2_{\mathsf{max}}(1+\discount)^2 \params\tr B\params.
            \end{align*}
        \end{lemma}
        \begin{proof}
        \noindent Note that
            \begin{align*}
                a_{j}\tr a_{j} &= \norm{\feature(\state_j)}_{2}^{2}\big[ \feature(\state_j)\feature(\state_j)\tr - \discount\{ \feature(\state_j) \feature(\state_{j}')\tr + \feature(\state_{j}')\feature(\state_j)\tr \} + \discount^{2} \feature(\state_{j}')\feature(\state_{j}')\tr
                \big].
            \end{align*}
        \noindent Therefore,
              \begin{align*}
                \params\tr\expecun{}[a_{j}\tr a_{j}]\params &= \params\tr\left(\norm{\feature(\state_j)}_{2}^{2}\expecun{}\big[ \feature(\state_j)\feature(\state_j)\tr - \discount\{ \feature(\state_j) \feature(\state_{j}')\tr + \feature(\state_{j}')\feature(\state_j)\tr \} + \discount^{2} \feature(\state_{j}')\feature(\state_{j}')\tr\big]\right)\params\\ 
                &\stackrel{(a)}{\leq} \params\tr\left(\Feature^2_{\mathsf{max}} \expecun{}\big[ \feature(\state_j)\feature(\state_j)\tr - \discount\{ \feature(\state_j) \feature(\state_{j}')\tr + \feature(\state_{j}')\feature(\state_j)\tr \} + \discount^{2} \feature(\state_{j}')\feature(\state_{j}'\right)\tr
                \big])\params\\
                &\stackrel{(b)}{\leq}\params\tr\left(\Feature^2_{\mathsf{max}} \expecun{}\big[ \feature(\state_j)\feature(\state_j)\tr - \discount\{ \feature(\state_j) \feature(\state_{j})\tr + \feature(\state_{j}')\feature(\state_j')\tr \} + \discount^{2} \feature(\state_{j}')\feature(\state_{j}')\tr
                \big]\right)\params\\
                &\leq \params\tr\left(\Feature^2_{\mathsf{max}} \left( \expecun{}[\feature(\state_j)\feature(\state_j)\tr + \discount^{2} \feature(\state_{j}')\feature(\state_{j}')\tr] - \discount \ \expecun{}[\feature(\state_j) \feature(\state_{j})\tr + \feature(\state_{j}')\feature(\state_j')\tr]\right)\right)\params\\
                &\stackrel{(c)}{\leq} \Feature^2_{\mathsf{max}}(1 + 2\discount + \discount^2) \params\tr B \params\\
                & = \Feature^2_{\mathsf{max}}(1+\discount)^2 \params\tr B \params,\\
        \end{align*}
        where $(a)$ follows from \Cref{asm:bddFeatures}, $(b)$ follows from \Cref{lem:psd-bound}, and $(c)$ holds because $\expecun{}^{\stationarydist}[\feature(\state_\timestep)\feature(\state_\timestep)\tr]  = \expecun{}^{\stationarydist}[\feature(\state'_\timestep)\feature(\state'_\timestep)\tr]$.
             
        \end{proof}

         \begin{lemma}\label{lem:high-prob-helper-1}
            Let $a_{j} \define [\feature(\state_{j})\feature(\state_{j}){\tr} - \discount\feature(\state_{j})\feature(\state'_{j}){\tr}]$. With $\lr \leq \lr_{\mathsf{max}} = \frac{1-\discount}{(1+\discount)^2\Phi^{2}_{\mathsf{max}}}$, the following bounds hold for any random variable $\params \in \real^d$ that is $\mathcal{F}_{j}$ measurable:
            \begin{align*}
                 \expecun{}\left[\params^{\top} \left(\identity - \lr a_j\right)\tr (\identity - \lr a_j)\params \vert \mathcal{F}_{j}\right] &\leq \left(1-\lr (1-\discount)\mu'\right)\norm{\params}_{2}^{2},\\
                 &\text{and}\\
                 \expecun{}\left[\norm{\left(\identity - \lr a_j\right)\params}_2 \vert \mathcal{F}_{j}\right] &\leq \left(1-\frac{\lr (1-\discount)\mu'}{2}\right)\norm{\params}_{2}.
            \end{align*}
        \end{lemma}
        \begin{proof}
            \noindent Note that 
            \begin{align*}
                \expecun{}[\params^{\top} (\identity - \lr a_j)\tr (\identity - \lr a_j)\params \vert \mathcal{F}_{j}] &= \expecun{}[\params^{\top} (\identity - \lr (a_{j}^{\top} + a_j) + \lr^2 a_j^{\top}a_j ) \params \vert \mathcal{F}_{j}]\\
                &= \norm{\params}^{2}_{2} - \lr \underbrace{\params^{\top} \expecun{}[ a^{\top}_j + a_j \vert\mathcal{F}_{j}] \params}_{\text{Term 1}} + \lr^2 \underbrace{\params^{\top}\expecun{}[ a_{j}^{\top}a_j\vert \mathcal{F}_{j}]\params}_{\text{Term 2}}.\numberthis\label{eq:hpb1}
            \end{align*}
            \noindent We now bound ``Term 1'' on the RHS above as follows:
            \begin{align*}
                 \params^{\top} \expecun{}[ a^{\top}_j + a_j \vert\mathcal{F}_{j}] \params &= \params^{\top} (A^{\top} + A)\params\\
                &\stackrel{(a)}{\geq}2(1-\discount)\params^{\top} B \params,\numberthis\label{eq:bound-tA1}
            \end{align*}
            where $(a)$ follows from \Cref{lem:A-lower}, with $B \define \expecun{}^{\stationarydist}[\feature(\state_\timestep)\feature(\state_\timestep)\tr]$.
            
           \noindent Next, we bound ``Term 2'' on the RHS of \cref{eq:hpb1} as follows:
            \begin{align*}
                \params^{\top}\expecun{}[ a_{j}^{\top}a_j\vert \mathcal{F}_{j}]\params \stackrel{(a)}{\leq}  \Feature^2_{\mathsf{max}}(1+\discount)^2 \params^{\top}B\params \numberthis\label{eq:bound-tA2},
            \end{align*}
            where $(a)$ is due to \Cref{lem:A-upper-2}.
            
            \noindent Compiling \eqref{eq:bound-tA1} and \eqref{eq:bound-tA2}, we get
            \begin{align*}
                \expecun{}[\params^{\top} (\identity - a_j)\tr (\identity - a_j)\params \vert \mathcal{F}_{j}] &\leq \norm{\params}_{2}^{2} - 2\lr(1-\discount)\params\tr B \params + \lr^{2}\Feature^{2}_{\mathsf{max}}(1+\discount)^{2} \params\tr B \params\\
                &= \norm{\params}_{2}^{2} -  \lr(1-\discount) \left(2 - \lr\frac{(1+\discount)^2}{(1-\discount)}\Feature^{2}_{\mathsf{max}}\right) \params\tr  B  \params\\
                &\stackrel{(a)}{\leq} \norm{\params}_{2}^{2}  - \lr (1-\discount) \params\tr  B  \params \\
                &\stackrel{(b)}{\leq} \norm{\params}_{2}^{2} -\lr (1-\discount)\mu' \norm{\params}_{2}^{2}\\
                &= (1-\lr\mu'(1-\discount))\norm{\params}^{2}_{2}\numberthis\label{eq:hpbound-abound-1},
            \end{align*}
            where $(a)$ uses $\lr \leq \lr_{\mathsf{max}}$, and $(b)$ follows from using Cauchy-Schwarz inequality along with the fact that $\mu'$ as the minimum eigenvalue of $B$.

            \noindent Applying Jensen's inequality in conjunction with \eqref{eq:hpbound-abound-1}, we obtain
            \begin{align*}
                \expecun{}[\norm{ (\identity - a_j)\params }_2 \vert \mathcal{F}_{j}] &\leq \left(1-\lr\mu'(1-\discount)\right)^{1/2}\norm{\params}_{2}\\
                &\stackrel{(a)}{\leq} \left(1-\frac{\lr\mu'(1-\discount)}{2}\right)\norm{\params}_2,
            \end{align*}
            where $(a)$ uses $1 - \lr \mu'(1-\discount) \geq 0 \implies (1 -  \lr \mu'(1-\discount))^{\frac{1}{2}} \leq 1 - \frac{ \lr \mu'(1-\discount)}{2}$.
        \end{proof}

    	\subsection{Expectation bound for Tail-averaged TD (Proof of \Cref{thm:expectation_bound})}	
    
        \subsubsection{Bias-variance decomposition of the non-asymptotic error} \label{sec:bias-var-decomp}

            Let,
            \begin{equation}
                C^{i:j}=
                \begin{cases}
                  (\identity - \lr a_i) (\identity - \lr a_{i-1})\ldots (\identity - \lr a_j) , & \text{if}\  i >= j \\
                  \identity, & \text{otherwise}
                \end{cases}
              \end{equation}
              where $a_\timestep$ is defined in \Cref{lem:A-upper-2}

            Next let, 
            \begin{align}
                z_{\timestep}^{\mathsf{bias}} &\define \expecun{}\left[\norm{C^{\timestep:1} z_0]}_{2}^{2}\right], \label{eq:z-bias-def}\\
                &\text{and}\\
                z_{\timestep}^{\mathsf{variance}} &\define \expecun{}\left[\norm{ \sum_{k=1}^{\timestep} C^{t:k+1}f_{k}(\params^\star)}_{2}^{2}\right].
                \label{eq:z-var-def}
            \end{align}
              
            \noindent Let the centered error be $z_{\timestep} = \params_{\timestep} - \params^{\star}$. Then, using \cref{eq:td-update}, we obtain 
            \begin{align*}
                z_{\timestep} &= \params_{\timestep-1} - \params^{\star}  + \lr (\cost_{\timestep} +  \discount\params^{\top}_{\timestep-1}\feature(\state'_{\timestep}) - \params^{\top}_{\timestep-1}\feature(\state_{\timestep}))\feature(\state_\timestep)\\
                &= \params_{\timestep-1} - \params^{\star} + \lr (\cost_\timestep\feature(\state_\timestep) - a_\timestep \params_{\timestep-1}) + \lr a_\timestep \params^{\star}  - \lr a_\timestep \params^{\star} \\
                &= (\identity - \lr a_\timestep)z_{\timestep-1} + \lr f_\timestep(\params^\star)\\
                &\stackrel{(a)}{=} C^{\timestep:1} z_0 + \lr \sum_{k=1}^{\timestep} C^{\timestep:k+1}f_{k}(\params^{\star}),\numberthis\label{eq:decomp}
            \end{align*}
            where $(a)$ follows from unrolling the update rule.

        \noindent Taking expectations on both sides of \eqref{eq:decomp}, we obtain
            \begin{align*}
                \expecun{}[ \norm{z_{\timestep}}_{2}^{2}] &= \expecun{}\left[\norm{C^{\timestep:1} z_0 + \lr \sum_{k=0}^{\timestep} C^{t:k+1}f_{k}(\params^\star)}_{2}^{2}\right]\\
                &\stackrel{(a)}{\leq}2z_{\timestep}^{\mathsf{bias}} + 2\lr^2 z_{\timestep}^{\mathsf{variance}},\numberthis\label{eq:bias-var-bound}
            \end{align*}
            where $(a)$ is obtained by using the following inequalities, i. $\norm{(a+b)}^{2}_2 \leq 2\norm{a}^{2}_{2} + 2\norm{b}^{2}_2$, ii. $\norm{\sum_{i=0}^{n} \ip_i} \leq \sum_{i=0}^{n}\norm{\ip_i}$. $z_{\timestep}^{\mathsf{bias}}$ and $z_{\timestep}^{\mathsf{variance}}$ are given by \eqref{eq:z-bias-def} and \eqref{eq:z-var-def}.

           \noindent Therefore, a bound on $\expecun{}\norm{z_{\timestep}}_{2}^{2}$ can obtained by bounding individual terms in \cref{eq:bias-var-bound}.
    
    \subsubsection{Bounding $z_{\timestep}^{\mathsf{bias}}$}  
    \begin{lemma}\label{lemma2}
        Under conditions of \Cref{thm:expectation_bound}, we have
        \begin{align}
             z_{\timestep}^{\mathsf{bias}} &\leq \exp(-\lr(1-\discount)\mu'\timestep)\expecun{}\left[\norm{z_{0}}^{2}_{2}\right], \ \forall t\ge 1.
        \end{align}
    \end{lemma}
    \begin{proof}
        Notice that
        \begin{align*}
            z_{\timestep}^{\mathsf{bias}} &= \expecun{}[\norm{C^{\timestep:1} z_0]}_{2}^{2}] \\
            &=\expecun{}\left[(C^{\timestep-1:1}z_{0})^{\top}(\identity - \lr a_\timestep)^{\top}(\identity - \lr a_\timestep)C^{\timestep-1:1}z_{0}\right]\\
            &\stackrel{(a)}{\leq}(1 - \lr(1-\discount)\mu')\expecun{}\left[\norm{C^{\timestep-1:1}z_{0}}_{2}^{2}\right]\\
            &\stackrel{(b)}{\leq} (1 - \lr(1-\discount)\mu')^{t}\expecun{}\left[\norm{z_{0}}^{2}_{2}\right]\numberthis\label{eq:bias-bound-res1}\nonumber\\
            &\leq \exp(- \lr(1-\discount)\mu'\timestep)\expecun{}\left[\norm{z_{0}}^{2}_{2}\right],\nonumber
        \end{align*}
        where $(a)$ follows from \Cref{lem:high-prob-helper-1}, and $(b)$ follows from using the argument in $(a)$ repeatedly.
    \end{proof}

    \subsubsection{Bounding $z_{\timestep}^{\mathsf{variance}}$}
        \begin{lemma}\label{lem:var-helper-1}
        For any $\mathcal{F}_{i}$ - measurable random vector $\ip \in \real^{d}$, and $\timestep \geq i$, we have
        
        \begin{align}
            \sum_{i = 0}^{\timestep}\expecun{}\left[\norm{C^{\timestep:i+1} \ip}^{2}_{2}\right] &\leq \frac{\expecun{}\left[\norm{\ip}^{2}_{2}\right]}{\lr(1-\discount)\mu'}.
        \end{align}
        \end{lemma}
        \begin{proof}
            \begin{align*}
            \expecun{}\left[\sum_{i = 0}^{\timestep}\norm{C^{\timestep:i+1} \ip}^{2}_{2}\right] &= \expecun{} \left[\sum_{i = 0}^{\timestep} \left(C^{\timestep:i+1} \ip\right)^{\top}
            \left(C^{\timestep:i+1} \ip)\right)\right] \\
            & = \expecun{} \left[\sum_{i = 0}^{\timestep} (C^{\timestep-1:i+1 }\ip)\tr(\identity-\lr a_i)\tr(\identity- \lr a_i)C^{\timestep-1:i+1}\ip\right]\\
            & \stackrel{(a)}{=} \sum_{i = 0}^{\timestep} \expecun{} \left[(C^{\timestep-1:i+1 }\ip)\tr(\identity-\lr a_i)\tr(\identity- \lr a_i)C^{\timestep-1:i+1}\ip\right]\\
            &\stackrel{(b)}{\leq} \sum_{i=0}^{\timestep}(1-\lr(1-\discount)\mu')\expecun{}\left[\norm{C^{\timestep-1:i+1}\ip}_{2}^{2}\right]\\
            &\stackrel{(c)}{\leq}\sum_{i=0}^{\timestep}(1-\lr(1-\discount)\mu')^{\timestep - i}\expecun{}\left[\norm{\ip}^{2}_{2}\right]\\
            &\stackrel{(d)}{\leq}\frac{\expecun{}[\norm{\ip}^{2}_{2}]}{\lr(1-\discount)\mu'},
        \end{align*} 
        where $(a)$ follows from linearity of the expectation operator, $(b)$ follows from \Cref{lem:high-prob-helper-1}, $(c)$ follows from unrolling the recursion, and $(d)$ follows from the fact that $\sum_{i=1}^{\timestep}(1- \lr (1-\discount)\mu')^{i} \leq \frac{1}{ \lr(1-\discount)\mu'}$.
            
        \end{proof}
        
        \begin{lemma}\label{lem:var-bound}
        Under conditions of Theorem \ref{thm:expectation_bound}, we have
         \begin{align}
             z_{\timestep}^{\mathsf{variance}} &\leq \frac{\sigma^{2}}{\lr(1-\discount)\mu'}.
         \end{align}
        \end{lemma}
        
        \begin{proof}
         
         \begin{align}
             z_{\timestep}^{\mathsf{variance}} &= 
             \sum_{i = 1}^{\timestep} \expecun{} \bigg[ \bigg( C^{\timestep:i+1} f_{i}(\params^\star)\bigg)^{\top} 
             \bigg( C^{\timestep:i+1} f_{i}(\params^\star)\bigg)\bigg]\\
             &\stackrel{(a)}{\leq}\frac{1}{\lr(1-\discount)\mu'}\expecun{}\left[\norm{f_{\timestep}(\params^\star)}_{2}^{2}| \mathcal{F}_{\timestep-1}\right]\\
             &\stackrel{(b)}{\leq}\frac{\sigma^{2}}{ \lr(1-\discount)\mu'},
        \end{align}  
        where $(a)$ follows from \Cref{lem:var-helper-1}, and $(b)$ follows from \Cref{asm:bddRewards}, and the fact that $\expecun{}\left[\norm{f_{\timestep}(\params^\star)}_{2}^{2}| \mathcal{F}_{\timestep-1}\right] \leq \sigma^2$, where $\sigma^2 = (\Cost_{\mathsf{max}} + (1+\discount)\Feature^{2}_{\mathsf{max}}\norm{\params^{\star}}^{2}_{2})$.
        \end{proof}

        \subsubsection{Proof of Theorem \ref{thm:expectation_bound}}
        \label{sec:expectation-bound}
            Recall that
            \begin{align*}
                 z_{k+1,N} &= \frac{1}{N}\sum_{i = k+1}^{k+N}z_{i}.
            \end{align*}
            \noindent Now,
            \begin{align}
                 \expecun{}\left[\norm{z_{k+1,N}}_{2}^{2}\right] &= \frac{1}{N^{2}}\sum_{i,j = k+1}^{k+N}\expecun{}\left[z_i^{\top} z_j\right]\nonumber\\
                                          &\stackrel{(a)}{\leq} \frac{1}{N^{2}}\bigg(\sum_{i= k+1}^{k+N}\expecun{}\left[\norm{z_i}_{2}^{2}\right] + 2 \sum_{i=k+1}^{k+N-1}\sum_{j=i+1}^{k+N} \expecun{}\left[z_{i}^{\top}z_{j}\right] \bigg)\numberthis\label{eq:cross-term-decomp},
            \end{align}
            where $(a)$ follows from separating out the diagonal and off-diagonal terms.
            
            \noindent For the bound in expectation for tail-averaged TD in \Cref{thm:expectation_bound}, we establish a useful result that bounds the second term in the RHS of \eqref{eq:cross-term-decomp}.
            
            \begin{lemma}\label{lem:tav-bias-support}
                For all $i\ge 1$, we have
                \begin{align}
                    \sum_{i=k+1}^{k+N-1}\sum_{j=i+1}^{k+N} \expecun{}\left[z_{i}^{\top}z_{j}\right]  
                    &\leq \frac{2}{\lr (1-\discount)\mu'}\sum_{i=k+1}^{k+N}\expecun{}\left[\norm{z_{i}}^{2}_{2}\right].
                \end{align}
            \end{lemma}
            \begin{proof}
                \begin{align*}
                   \sum_{i=k+1}^{k+N-1}\sum_{j=i+1}^{k+N} \expecun{} \left[z_{i}^{\top}z_{j}\right]  &=  
                   \sum_{i=k+1}^{k+N-1}\sum_{j=i+1}^{k+N} \expecun{} \left[z_{i}^{\top}(C^{(j:i+1)}z_{i} + \lr\sum_{l=0}^{j-i-1} C^{j:l+1}f_{l}(\params^{\star})))\right]\\
                   &\stackrel{(a)}{=}\sum_{i=k+1}^{k+N-1}\sum_{j=i+1}^{k+N} \expecun{}  \left[z_{i}^{\top}C^{(j:i+1)}z_{i}\right] \\
                   &=\sum_{i=k+1}^{k+N-1}\sum_{j=i+1}^{k+N} \expecun{}  \left[z_{i}^{\top}(\identity - \lr a_j)(C^{(j-1:i+1)} z_{i})\right] \\
                   &\stackrel{(b)}{\leq} \sum_{i=k+1}^{k+N-1}\sum_{j=i+1}^{k+N} \expecun{} \left[ \norm{z_{i}}_{2}^{2} \norm{C^{j:i+1}}_{2} \right]\\
                   &\stackrel{(c)}{\leq}\sum_{i=k+1}^{k+N-1}\sum_{j=i+1}^{k+N} \left(1 - \frac{\lr (1-\discount)\mu'}{2}\right)^{j-i}\expecun{}\left[\norm{z_{i}}^{2}_{2}\right]\\
                   &\leq \sum_{i=k+1}^{k+N}\expecun{}\left[\norm{z_{i}}^{2}_{2}\right]\sum_{j=i+1}^{\infty}\left(1- \frac{\lr (1-\discount)\mu'}{2}\right)^{j-i}\\
                   &\stackrel{(d)}{\leq}\frac{2}{\lr (1-\discount)\mu'}\sum_{i=k+1}^{k+N}\expecun{}\left[\norm{z_{i}}^{2}_{2}\right],
                \end{align*}
                
            where $(a)$ follows from the fact that $\expecun{}[f_\timestep(\params^\star)\vert \mathcal{F}_{\timestep-1}] = 0$, $(b)$ follows from the fact that $z\tr C^{j:i+1} z \leq \norm{z}_{2}^{2}\norm{C^{j:i+1}}_2$ due to Cauchy-Schwarz inequality, $(c)$ is because $ \norm{C^{j:i+1}}_{2}\leq \left (\norm{ I - \lr a}_2 \right)^{j-i} \leq \left( 1 - \frac{\lr(1-\discount) \mu'}{2}\right)^{j-i}$ (due to Cauchy-Schwarz inequality and \Cref{lem:high-prob-helper-1}), and $(d)$ follows from the summation of the geometric series.
            \end{proof}
                
            \noindent For the sake of readability, we will restate Theorem \ref{thm:expectation_bound} below.
             \begin{theorem}
             Suppose  \Crefrange{asm:stationary}{asm:phiFullRank} hold.  Choose a step size $\lr$ satisfying
                \begin{align*}
                \lr \leq \lr_{\mathsf{max}} =  \frac{1-\discount}{(1+\discount)^2\Phi^{2}_{\mathsf{max}}},
                \numberthis\label{eq:gammamax-appd}
                \end{align*}
                where $\beta$ is the discount factor and $\Phi_{\mathsf{max}}$ is a bound on the features (see \Cref{asm:bddFeatures}).
                
                Then the expected error of the tail-averaged iterate $\params_{k+1,N}$ formed using Algorithm \ref{alg:ciac-a} satisfies
            \begin{align*}
               \expecun{}\left[\norm{\params_{k+1,N} - \params^{\star}}^{2}_{2}\right] &\leq \frac{10  e^{(-k \lr (1-\discount)\mu')}}{\lr^2 (1-\discount)^2\mu'{}^{2} N^{2}}\expecun{}\left[\norm{\params_{0} - \params^{\star}}^{2}_2\right] + \frac{10\sigma^{2}}{(1-\discount)^2\mu'{}^2 N},\numberthis\label{eq:td-expec-bd-appd}
            \end{align*}
               where $N = \timestep - k$, $\sigma^2 = (\Cost_{\mathsf{max}} + (1+\discount)\Feature^{2}_{\mathsf{max}}\norm{\params^{\star}}_{2}^{2})$, $\theta_0$ is the initial point, and $\theta^\star$ is the TD fixed point specified in \eqref{eq:theta-star}.
              \end{theorem}    
            \begin{proof} \label{proof:thm-1}
                Substituting the result of \Cref{lem:tav-bias-support} in \cref{eq:cross-term-decomp}, we obtain
                \begin{align*}
                     \expecun{}\left[\norm{z_{k+1,N}}_{2}^{2}\right] &\leq \frac{1}{N^{2}}\left(\sum_{i = k+1}^{k+N}\expecun{}\left[\norm{z_i}_{2}^{2}\right] + \frac{4}{\lr (1-\discount)\mu'} \sum_{i = k+1}^{k+N}\expecun{}\left[\norm{z_i}_{2}^{2}\right] \right)\\
                     &= \frac{1}{N^{2}}\left(1+\frac{4}{\lr (1-\discount)\mu'}\right) \sum_{i = k+1}^{k+N}\expecun{}\left[\norm{z_i}_{2}^{2}\right]\\
                     &\stackrel{(a)}{\leq} \underbrace{\frac{2}{N^{2}}\left(1+\frac{4}{\lr (1-\discount)\mu'}\right) \sum_{i = k+1}^{k+N}   z_{i}^{\mathsf{bias}}}_{z_{k+1, N}^{\mathsf{bias}}} + \underbrace{\frac{2}{N^{2}}\bigg(1+\frac{4}{\lr (1-\discount)\mu'}\bigg)\lr^{2}  \sum_{i = k+1}^{k+N}   z_{i}^{\mathsf{variance}}}_{z_{k+1,N}^{\mathsf{variance}}}\numberthis\label{eq:tail-av-bias-var-2}.
                \end{align*}
                where $(a)$ follows from \cref{eq:bias-var-bound}.
                
                \noindent $z_{k+1,N}^{\mathsf{bias}}$in \cref{eq:tail-av-bias-var-2}  is bounded as follows:
                \begin{align*}
                            z_{k+1,N}^{\mathsf{bias}} &\leq \frac{2}{N^{2}}\left(1 + \frac{4}{\lr (1-\discount)\mu'}\right) \sum_{i = k+1}^{\infty}   z_{i}^{\mathsf{bias}} \\
                            &\stackrel{(a)}{\leq} \frac{2}{N^{2}}\left(1 + \frac{4}{\lr (1-\discount)\mu'}\right) \sum_{i = k+1}^{\infty} (1 - \lr (1-\discount)\mu')^{i}\expecun{}\left[\norm{z_{0}}^{2}_{2}\right] \\
                            &\stackrel{(b)}{=}\frac{2}{\lr (1-\discount)\mu' N^{2}}\left(1-\lr(1-\discount)\mu'\right)^{k+1}\left(1 + \frac{4}{\lr (1-\discount)\mu'}\right)\expecun{}\left[\norm{z_{0}}^{2}_{2}\right],
                        \end{align*}
                   where $(a)$ follows from \cref{eq:bias-bound-res1} in proof of \Cref{lemma2}, and $(b)$ follows from the bound on summation of a geometric series.
                   
                $z_{k+1,N}^{\mathsf{variance}}$ in \cref{eq:tail-av-bias-var-2} is bounded as follows:
                \begin{align*}
                        z_{k+1,N}^{\mathsf{variance}}
                        &\stackrel{(a)}{\leq}\frac{2\lr^2}{N^{2}}\left(1+\frac{4}{\lr (1-\discount)\mu'}\right)\sum_{i=k+1}^{k+N}\frac{\sigma^2}{\lr(1-\discount)\mu'}\\
                        &\leq \frac{2\lr^2}{N^{2}}\left(1+\frac{4}{\lr (1-\discount)\mu'}\right)\sum_{i=0}^{N}\frac{\sigma^2}{\lr(1-\discount)\mu'}\\
                        &= \bigg(1+\frac{4}{\lr (1-\discount)\mu'}\bigg)\frac{2\lr\sigma^2}{(1-\discount)\mu' N},
                    \end{align*}
                    where  $(a)$ follows from \Cref{lem:var-bound}.
                
              Finally substituting the bounds on $z_{k+1,N}^{\mathsf{bias}}$ and $z_{k+1,N}^{\mathsf{variance}}$ in \eqref{eq:tail-av-bias-var-2}, we get
            \begin{align*}
                \expecun{}[\norm{z_{k+1,N}}^{2}_{2}] &\leq\bigg(1+ \frac{4}{\lr(1-\discount)\mu'}\bigg)\bigg(\frac{2}{\lr (1-\discount)\mu' N^{2}}(1-\lr(1-\discount)\mu')^{k+1}\expecun{}[\norm{z_{0}}^{2}_{2}] + \frac{2\lr\sigma^2}{(1-\discount)\mu' N}\bigg),\\
                &\stackrel{(a)}{\leq}\bigg(1+ \frac{4}{\lr(1-\discount)\mu'}\bigg)\bigg(\frac{2 \exp(-k \lr (1-\discount)\mu')}{\lr (1-\discount)\mu' N^{2}}\expecun{}[\norm{z_{0}}^{2}_{2}] + \frac{2\lr\sigma^2}{(1-\discount)\mu' N}\bigg)\\
                &\stackrel{(b)}{\leq}\frac{10 \exp(-k \lr (1-\discount)\mu')}{\lr^2 (1-\discount)^2\mu'{}^2 N^{2}}\expecun{}\left[\norm{z_{0}}^{2}_{2}\right] + 
                \frac{10\sigma^2}{(1-\discount)^2\mu'{}^2 N},
            \end{align*}
                where $(a)$ follows from the fact that $(1+x)^{y} = \exp(y\log(1+x)) \leq \exp(xy)$, and $(b)$ uses $\lr(1-\discount)\mu' < 1$ as $\mu' < \Feature_{\mathsf{max}}^2$, which implies that $1+ \frac{4}{\lr(1-\discount)\mu'} \leq \frac{5}{\lr(1-\discount)\mu'}$.
            \end{proof}
            
    \subsection{High probability bound for tail-averaged TD (Proof of \Cref{thm:high-prob-bound})} \label{sec:high-prob}

    \begin{proposition}\label{prop:hprob}
        Suppose \Cref{asm:bddFeatures,asm:bddRewards} hold, then for all $\epsilon \geq 0$, and $\timestep \geq 1$,
        \begin{align*}
            \prob\left(\norm{z_{k+1,N}}_{2} - \expecun{}\left[\norm{z_{k+1,N}}_{2} > \epsilon\right]\right) \leq \exp\left(-\frac{\epsilon^{2}}{ (\Cost_{\mathsf{max}} + (1+ \discount) H \Feature^{2}_{\mathsf{max}})^{2} \sum_{i = k+1}^{k+N} L^{2}_{i}} \right),
        \end{align*}
    where $L_{i} \define \frac{\lr}{N}\sum_{j=i+1}^{i+N}\left(1 - \frac{\lr(1-\discount)\mu'}{2}\right)^{j-i+1}$.
    \end{proposition}
    \begin{proof}
        To derive the result we follow the technique from \citep{lstd-prashanth}. 
        
        \noindent\textbf{Step 1.} We decompose the centered error $\norm{z_{k+1, N}}_{2}  - \expecun{}[\norm{z_{k+1,N}}_{2}]$ as follows:
        \begin{align*} \numberthis\label{eq:D}
            \norm{z_{k+1, N}}_{2} - \expecun{}\left[\norm{z_{k+1,N}}_{2}\right]
            &=\sum_{i=k+1}^{k+N}D_{i},
        \end{align*}
        where $D_i \define g_i -\expecun{}\left[ g_i\vert \mathcal{F}_{i-1}\right]$, and $g_i \define \expecun{}[ \norm{z_{k+1,N}}_{2} \vert \mathcal{F}_{i-1}]$.

       \noindent \textbf{Step 2.}

        \noindent We prove that functions $g_{i}$ are Lipschitz continuous in the random innovation $f_i$ (given by \eqref{eq:f}) at time $i$ with constant $L_{i}$.

        \noindent First, let $\Theta^{i}_{k+1,N}(\params)$ to be the value of the tail-averaged iterate at time $\timestep$ that evolves according to \cref{eq:td-update} beginning from $\params$ at time $i$. Next, let $\bar{\Theta}^{i}_{k+1,N}(\bar{\params}, \params )$ be defined as follows:
    
        \begin{align*}
            \bar{\Theta}^{i}_{k+1,N}(\bar{\params}, \params ) &\define\frac{(i-k)\bar{\params}}{N} + \frac{1}{N}\sum_{j = i+1}^{i+N}\Theta^{i}_{j}(\params),
        \end{align*}
        where $\bar\params$ is the value of the tail averaged iterate at time $i$.

        \noindent Now let $f$ and $f'$ denote two possible values of the random innovation at time $i$, and set $\params = \params_{i-1} + \lr f$ and $\params' = \params_{i-1} + \lr f'$. Therefore,
        
        \begin{align*}
            \expecun{}\left[\norm{\bar{\Theta}^{i}_{i+1,N}(\bar{\params}, \params) -  \bar{\Theta}^{i}_{k+1,N}(\bar{\params}, \params') }_{2}\right] &=
            \expecun{}\bigg[ \frac{1}{N}\sum_{j = i+1}^{i+N}\norm{ (\Theta^{i}_{j}(\params) - \Theta^{i}_{j}(\params'))}_{2}\bigg].\numberthis\label{eq:interm}
        \end{align*}
        
        \noindent We will now bound the term $\Theta^{i}_{j}(\params) - \Theta^{i}_{j}(\params')$ inside the summation of \eqref{eq:interm}.

        We will first , note that as the projection $\Gamma$ is non-expansive, we have the following
        \begin{align*}
            \expecun{}\bigg[\norm{\Theta_{j}^{i}(\params) - \Theta_{j}^{i}(\params')}_{2}|\mathcal{F}_{j-1}\bigg]
            &\leq \expecun{}\bigg[ \norm{\Theta_{j-1}^{i}(\params) - \Theta_{j-1}^{i}(\params') - \lr [f_{i}(\Theta_{j-1}^{i}(\params)) - f_{i}(\Theta_{j-1}^{i}(\params')])}_{2}|\mathcal{F}_{j-1}\bigg].\numberthis\label{eq:high-prob-td-decomp}
        \end{align*}
          
        \noindent Expanding $f_i$ and using the definition $a_{j}$ from \Cref{lem:A-upper-2}, we have
         \begin{align*}
            \nonumber\Theta_{j-1}^{i}(\params) &- \Theta_{j}^{i}(\params') - \lr [f_{i}(\Theta_{j-1}^{i}(\params)) - f_{i}(\Theta_{j-1}^{i}(\params')]\\
             &= \Theta_{j-1}^{i}(\params) - \Theta^{i}_{j-1}(\params') - \lr [\feature(\state_{j})\feature(\state_{j}){\tr} - \discount\feature(\state_{j})\feature(\state'_{j}){\tr} ] [(\Theta^{j}_{i-1}(\params)) - (\Theta^{j}_{i-1}(\params')]\\
             &= [\identity - \lr a_{j}](\Theta_{j-1}^{i}(\params) - \Theta_{j-1}^{i}(\params')). 
        \end{align*}

        \noindent Using the tower property of conditional expectations, it follows that
        \begin{align*}
            \expecun{}[\norm{\Theta^{i}_{j}(\params) - \Theta^{i}_{j}(\params')}_{2} ] &= \expecun{}\bigg[\expecun{}\left[\Vert \Theta^{i}_{j}(\params) - \Theta^{i}_{j}(\params')\Vert_{2} \bigg \vert \mathcal{F}_{j-1}\right]\bigg]\\
            &= \expecun{}\bigg[ \expecun{}\left[\bigg \Vert (\identity - \lr a_j)(\Theta_{j-1}^{i}(\params) - \Theta_{j-1}^{i}(\params'))\bigg \Vert_{2} \bigg \vert \mathcal{F}_{j-1}\right]\bigg]\\
            &\stackrel{(a)}{\leq} \left(1-\frac{ \lr(1-\discount)\mu'}{2}\right) \expecun{}[\norm{\Theta^{i}_{j-1}(\params) - \Theta^{i}_{j-1}(\params')}_{2}]\\
            &\stackrel{(b)}{=} \left(1 - \frac{\lr(1-\discount)\mu'}{2}\right)^{j-i+1} \norm{\params - \params'}_{2},
        \end{align*}
        where $(a)$ follows from \Cref{lem:high-prob-helper-1}, and $(b)$ follows from repeated application of argument used in arriving at $(a)$. 
        
       Now, using Jensen's inequality,  we get
        
        \begin{align*}
            \vert \expecun{}[\norm{ \params_{j} - \params^\star}_{2}\vert \params_i = \params] - \expecun{}[\norm{ \params_{j} - \params^\star}_{2}\vert \params_i = \params'] \vert &\leq \expecun{}\bigg[\norm{ (\Theta^{i}_{j}(\params) - \Theta^{i}_{j}(\params'))}_{2}\bigg],\\
            &\leq \lr \left(1 - \frac{\lr(1-\discount) \mu'}{2}\right)^{j-i+1} \norm{f - f'}_{2}. \numberthis\label{eq:lip-comp}
        \end{align*}
        
        \noindent Substituting \eqref{eq:lip-comp} in \eqref{eq:interm}, we obtain
        \begin{align*}
            \expecun{}[\norm{\bar{\Theta}^{i}_{k+1}(\bar{\params}, \params) -  \bar{\Theta}^{i}_{k+1}(\bar{\params}, \params') }_{2}] 
            &\leq \frac{\lr}{N}  \sum_{j=i+1}^{i+N}\left(1 - \frac{\lr (1-\discount)\mu'}{2}\right)^{j-i+1} \norm{f - f'}_{2}. \numberthis\label{eq:lip-up}
        \end{align*}
        
        \noindent From \eqref{eq:lip-comp}, it is clear that $g_{i}$ is $L_{i}$- Lipschtiz in the random innovation at time $i$, which implies that $D_{i}$ is Lipschitz with the constant $L_{i}$.

    
    \noindent {\bf Step 3.}

       \noindent Next, we derive a standard martingale concentration bound for the tail-averaged iterate $z_{k+1, N}$. For any $\eta > 0$,
        \begin{align*}
            \prob(\norm{z_{k+1, N}}_{2}^{2} - \expecun{}[\norm{z_{k+1, N}}_{2}^{2}] \geq \epsilon) &= \prob \bigg( \sum_{i = k+1}^{k+N} D_{i} \geq \epsilon \bigg) \\
            &\stackrel{(a)}{\leq} \exp(-\eta \epsilon) \expecun{}\bigg[ \exp( \eta  \sum_{i = k+1}^{k+N} D_{i})\bigg] \\
            &\stackrel{(b)}{=} \exp(-\eta \epsilon) \expecun{}\bigg[ \exp( \eta  \sum_{i = k+1}^{k+N} D_{i}) \expecun{}\bigg[ \exp(\eta D_{k+1,N})| \mathcal{F}_{\timestep-1}\bigg]\bigg],
        \end{align*}
        where are $(a)$ follows from Markov inequality and $(b)$ follows from \cref{eq:D}.
        
         \noindent From \citep[Proof of proposition 1 part 3, page 585]{lstd-prashanth}, we have the following bound for a zero-mean r.v. $Z$ with $|Z| \leq B$ w.p 1, and a $L$-Lipschitz function $g$:
        \begin{align*}
            \expecun{}[\exp(\eta g(Z))] 
            &\leq \exp\bigg(\frac{\eta^{2}B^{2}L^{2}}{2}\bigg).
        \end{align*} 
        Next, from \cref{asm:bddRewards}, and the projection step of the algorithm we have that $f_{i}(\params_{i-1}) < (\Cost_{\mathsf{max}} + (1+ \discount) H \Feature^{2}_{\mathsf{max}})$ is a bounded random variable, and conditioned on $\mathcal{F}_{i-1}$ , $D_{i}$ is Lipshitz in $f_{i}(\params_{i-1})$ with constant $L_{i}$. Hence,
        \begin{align*}
            \expecun{}[\exp(\eta D_{\timestep})| \mathcal{F}_{\timestep-1}] &\leq \exp\bigg( \frac{ \eta^{2}(\Cost_{\mathsf{max}} + (1+ \discount) H \Feature^{2}_{\mathsf{max}})^{2} L_{\timestep}^{2}}{2}\bigg).
        \end{align*}
        
        Using these facts we get,
        \begin{align*}
            \prob\left(\norm{z_{k+1, N}}_{2} - \expecun{}\left[\norm{z_{k+1, N}}_{2}\right] > \epsilon\right) &\leq \exp(-\eta \epsilon) \exp\bigg( \frac{ \eta^{2}(\Cost_{\mathsf{max}} + (1+ \discount) H \Feature^{2}_{\mathsf{max}})^{2} \sum_{i=k+1}^{k+N}L_{i}^{2}}{2}\bigg).
        \end{align*}
Optimising over $\eta$ in the above inequality leads to
        \begin{align*}
            \prob\left(\norm{z_{k+1,N}}_{2} - \expecun{}\left[\norm{z_{k+1,N}}_{2}\right] > \epsilon \right) &\leq \exp\bigg(-\frac{\epsilon^{2}}{ (\Cost_{\mathsf{max}} + (1+ \discount) H \Feature^{2}_{\mathsf{max}})^{2} \sum_{i = k+1}^{k+N} L^{2}_{i}} \bigg).
        \end{align*}
    \end{proof}

   \subsubsection{Bounding the Lipschitz constant}
   
    \begin{lemma}\label{lem:lipschitz}
        With $L_i$ as defined in \Cref{prop:hprob}, we have
        \begin{align*}
            \sum_{i=k+1}^{k+N} L_{i}^{2} &\leq \frac{4}{N(1-\discount)^2\mu'^2}.
        \end{align*}
    \end{lemma}
    \begin{proof} Notice that
         \begin{align*}
             \sum_{i=k+1}^{k+N} L_{i}^{2} &=\frac{\lr^2}{N^2}\sum_{i=k+1}^{k+N}\bigg(\sum_{j=i+1}^{i+N}\left(1 - \frac{\lr (1-\discount)\mu'}{2}\right)^{j-i+1}\bigg)^2\\
            &\leq \frac{\lr^2}{N^2}\sum_{i=k+1}^{k+N}\bigg(\sum_{j=i+1}^{\infty}\left(1 - \frac{\lr (1-\discount)\mu'}{2}\right)^{j-i+1}\bigg)^2\\
            &=\frac{\lr^2}{N^2}\sum_{i=k+1}^{k+N}\left(\frac{4}{\lr^2 (1-\discount)^2\mu'^2}\right)\\
            &= \frac{4}{N(1-\discount)^2\mu'^2}.
        \end{align*}
    \end{proof}
   
	\subsubsection{Proof of Theorem \ref{thm:high-prob-bound}}   
   \noindent For the sake of convenience,  we restate the high probability bound for tail-averaged TD below.
    \begin{theorem}[\textbf{High-probability bound}]\label{thm:high-prob-bound-2}
    	Suppose \Crefrange{asm:stationary}{asm:projection} hold. Choose the step size such that $\lr \leq \lr_{\mathsf{max}}$, where $\lr_{\mathsf{max}}$ is defined in \eqref{eq:gammamax}. Then, for any  $\delta\in (0,1]$, we have the following bound for the projected tail-averaged iterate $\params_{k+1,N}$:
    \begin{align*}
        \prob\left(\norm{\params_{k+1,N} - \params^\star}_{2} \leq \frac{2\sigma}{(1-\discount)\mu' \sqrt{N}}\sqrt{\log\left(\frac{1}{\delta}\right)} +  \frac{\sqrt{10} e^{(-k \lr (1-\discount)\mu')}}{\lr (1-\discount)\mu'N}\expecun{}\left[\norm{\params_{0} - \params^\star}_{2}\right] + \frac{\sqrt{10}\sigma}{(1-\discount)\mu' \sqrt{N}} \right) &\geq 1 - \delta,
    \end{align*}
    where $N,\sigma,\mu, \theta_0,\theta^\star$ are as specified in Theorem \ref{thm:expectation_bound}.
    \end{theorem}
    \begin{proof}\label{proof:hpb-1}
     \begin{align*}
            \prob(\norm{z_{k+1,N}}_{2}^{2} - \expecun{}[\norm{z_{k+1,N}}_{2}^{2}]> \epsilon ) &\stackrel{(a)}{\leq} \exp\bigg(-\frac{\epsilon^{2}}{ (\Cost_{\mathsf{max}} + (1+ \discount) H \Feature^{2}_{\mathsf{max}})^{2} \sum_{i = k+1}^{k+N} L^{2}_{i}} \bigg)\\
            &\stackrel{(b)}{\leq} \exp\bigg(-\frac{N(1-\discount)\mu'^2 \epsilon^2}{4\sigma^2}\bigg)\numberthis\label{eq:crude_hprob},
        \end{align*}
    where $(a)$ follows from \Cref{prop:hprob}, and $(b)$ follows from \Cref{lem:lipschitz}.
    
    \noindent The inequality in \cref{eq:crude_hprob} can be re-written in high-confidence form as follows: For any $\delta \in (0,1]$,
    \begin{align*}
            \prob\left(\norm{z_{k+1,N}}_{2}^{2} - \expecun{}[\norm{z_{k+1,N}}_{2}^{2}]\leq \frac{2\sigma}{(1-\discount)\mu'\sqrt{N}}\sqrt{\log\left(\frac{1}{\delta}\right)} \right) \geq  1-\delta.
        \end{align*}
    \noindent The final bound follows by substituting the bound on $\expecun{}\left[\norm{z_{k+1,N}}_{2}^{2}\right]$ from  \Cref{thm:expectation_bound} in the inequality above.
    \end{proof}

   \subsection{Expectation bound for Tail-averaged TD with regularisation}    
   
        We will first begin by rewriting the regularised TD iteration as follows:
    
        \begin{align*}
            \hat\params_{\timestep+1} &= (\identity - \lr \lambda)\hat\params_{\timestep} + \lr (-A \hat\params_\timestep + b \Delta M_\timestep),
        \end{align*}
        where $ \Delta M_\timestep = \hat f_{t}(\hat{\params}_{t-1}) - \expecun{}^{\rho,P}\left[ \hat f_{t}(\hat{\params}_{t-1}) \vert \mathcal{F}_{\timestep -1}\right]$.

        \noindent Recall that the solution for the regularised TD iteration is given by
        \begin{align*}
            \params^{\star}_{\mathsf{reg}} &= (A + \lambda \identity)^{-1}b,
        \end{align*}
        where $A$ and $b$ are defined in \cref{eq:theta-star}.
        
         \begin{lemma}\label{lem:regtd-featurebound}
            With $\lr \leq \lr_{\mathsf{max}} = \frac{\lambda}{\lambda^2 + 2\lambda (1+\discount)\Phi^{2}_{\mathsf{max}} + (1+ \discount)^2 \Phi^4_{\mathsf{max}}}$, 
            we have
            \begin{align*}
                 \bigg\Vert\bigg(\identity - \lr(A + \lambda\identity)\bigg)^{\top} \bigg(\identity - \lr(A + \lambda\identity)\bigg)\bigg\Vert_2 &\leq 1 - \lr(\mu+\lambda),\\
                 &\text{and}\\
                 \Vert(\identity - \lr A)\Vert_2 &\leq 1 - \frac{\lr(\mu+\lambda)}{2}.
            \end{align*}
            \end{lemma}
            \begin{proof}
                \begin{align*}
                    \bigg\Vert(\identity - \lr (A + \lambda \identity))\tr (\identity - \lr (A + \lambda \identity))\bigg\Vert_2 &= \bigg\Vert\identity - \lr (A + \lambda \identity) - \lr (A + \lambda \identity)\tr + \lr^2 (A + \lambda \identity)\tr (A + \lambda \identity)\bigg\Vert_2\\
                    &= \bigg\Vert\identity - \lr ( A + A\tr + 2\lambda \identity) +  \lr^2 (A + \lambda \identity)\tr (A + \lambda \identity)\bigg\Vert_2\\
                    &\leq \bigg\Vert \identity - \lr ( A + A\tr + 2\lambda \identity)\bigg\Vert_2 + \bigg\Vert \lr^2 (A + \lambda \identity)\tr (A + \lambda \identity) \bigg\Vert_2\\
                    &= \bigg\Vert \identity - \lr ( A + A\tr + 2\lambda \identity)\bigg\Vert_2 + \lr^2 \bigg \Vert  A\tr A + \lambda (A\tr + A) + \lambda^2 \identity \bigg\Vert_2\\
                    &\stackrel{(a)}{\leq} 1 - \lr\big(2 \mu + 2\lambda - \lr(\lambda^2 + 2\lambda (1+\discount)\Phi^{2}_{\mathsf{max}} + (1+ \discount)^2 \Phi^4_{\mathsf{max}})\big)\\
                    &\stackrel{(b)}{\leq} 1 - (2\mu+ \lambda)\lr\\
                    &\leq 1 - \lr(\mu + \lambda),
                \end{align*}
               where (a) follows from \Cref{lem:A-upper}, and $\Vert(A + A\tr)\Vert_2^{2} \geq 2\mu $  and $(b)$ follows as per definition of $\lr$.

               \noindent For the second claim, note that $1 - \lr (\mu+\lambda) \geq 0 \implies (1 - \lr (\mu+ \lambda))^{\frac{1}{2}} \leq 1 - \frac{\lr(\mu+\lambda)}{2}$. Therefore,
                \begin{align*}
                    \Vert(\identity - \lr A)\Vert_2 \leq 1 - \frac{\lr(\mu+\lambda)}{2}.
                \end{align*}
            \end{proof}

            \begin{lemma}\label{lem:regtd-highprob-helper1}
                Let $\lr \leq \lr_{\mathsf{max}}$, where $\lr_{\mathsf{max}}$ is set as per \eqref{eq:reg-td-lr}, and $a_{j}$ be as defined in \Cref{lem:A-upper-2}. Then for any $\mathcal{F}_{i-1}$ measurable $\hat \params \in \real^d$ we have 
                \begin{align*}
                    \expecun{}\left[\hat{\params}^{\top} (\identity - \lr (\lambda \identity + a_j))\tr (\identity - \lr (\lambda \identity + a_j))\hat{\params} \vert \mathcal{F}_{j}\right] &\leq \left(1-\lr(\mu+\lambda)\right)\norm{\hat \params}_{2}^{2}\\
                    &\text{and,}\\
                    \expecun{}\left[\norm{(\identity - \lr(\lambda \identity+a_j))\hat{\params}}_{2}\vert\mathcal{F}_{j}\right] &\leq \left(1-\frac{\lr(\mu+\lambda)}{2}\right)\norm{\hat \params}_{2}.
                \end{align*}
            \end{lemma}
            \begin{proof}
                 \begin{align*}
                    \expecun{}[\hat{\params}^{\top} (\identity - \lr (\lambda \identity & + a_j))\tr (\identity - \lr (\lambda \identity + a_j))\hat{\params} \vert \mathcal{F}_{j}]\\ &= \expecun{}[\hat{\params}^{\top} (\identity - 2\lr\lambda\identity - \lr(a_j + a_{j}\tr) + \lr^2\{ \lambda^2 \identity + \lambda  (a_j + a_{j}\tr) + a_{j}\tr a_j\} ) \hat{\params} \vert \mathcal{F}_{j}]\\
                    &= \expecun{}[\hat{\params}^{\top}\hat{\params}\vert\mathcal{F}_{j}] -  \lr \expecun{}[\hat{\params}\tr 2\lambda \identity \hat{\params} \vert \mathcal{F}_{j}] -\lr \underbrace{\hat{\params}^{\top} \expecun{}[ a^{\top}_j + a_j \vert\mathcal{F}_{j}] \hat{\params}}_{\text{Term 1}} +\\ 
                    &\lr^2 \underbrace{\hat{\params}^{\top}\expecun{}[ a_{j}^{\top}a_j\vert \mathcal{F}_{j}]\hat{\params}}_{\text{Term 2}} + \lr^2 \underbrace{\hat{\params}\tr\expecun{}[a_{j} + a_{j}\tr\vert \mathcal{F}_{j}]\hat{\params} }_{\text{Term 3}}+ \lr^2 \expecun{}[\hat{\params}\tr \lambda^2\identity \hat{\params} \vert \mathcal{F}_{j}]  .
                \end{align*}
                We proceed by bounding the Terms 1, 2, and 3 individually.
                
                \noindent Term 1:
        
                \begin{align*}
                     \params^{\top} \expecun{}[ a^{\top}_j + a_j \vert\mathcal{F}_{j}] \params &= \params^{\top} (A^{\top} + A)\params\\
                    &\stackrel{(a)}{\geq}2\mu\norm{\hat{\params}}_{2}^{2},\numberthis\label{eq:bound-A1-regtd}
                \end{align*}
                where, $\mu$ is the minimum eigenvalue of matrix $A$.\\[1ex]
                \noindent Term 2:
                
                \noindent From \Cref{lem:A-upper-2} for any $\hat{\params} \in \real^{d}$, we have 
                \begin{align*}
                    \params\tr(\expecun{}[a_{j}\tr a_{j}])\params &\leq \Feature^2_{\mathsf{max}}(1+\discount)^2 \hat{\params}\tr B\hat{\params}\\
                                                                  &\stackrel{(a)}{\leq}\Feature^2_{\mathsf{max}}(1+\discount)^2 \norm{\hat\params}_{2}^{2}\norm{B}_{2}\\
                                                                  &\stackrel{(b)}{\leq} (1+\discount)^{2}\Feature^{4}_{\mathsf{max}}\norm{\hat\params}_{2}^{2}, 
                \end{align*}
                where $(a)$ follows from the Cauchy-Schwarz inequality, and $(b)$ follows from \Cref{asm:bddFeatures}.\\[1ex]
               \noindent Term 3:
               \begin{align*}
                   \hat{\params}\tr\expecun{}[a_{j} + a_{j}\tr\vert \mathcal{F}_{j}]\hat{\params} &= \hat{\params}\tr (A+ A\tr) \hat{\params}\\
                   &\stackrel{(a)}{\leq} \norm{\hat{\params}}_{2}^{2}\norm{A + A\tr}_{2}\\
                   &\stackrel{(b)}{\leq} 2\norm{\hat{\params}}_{2}^{2}(1+\discount)\Feature^{2}_{\mathsf{max}},
               \end{align*}
               where $(a)$ follows from Cauchy-Schwarz inequality and $(b)$ follows from \Cref{lem:A-upper}.

               \noindent Combining the bounds on Term 1,2, and 3, we obtain
               \begin{align*}
                   \expecun{}[\hat{\params}^{\top} (\identity - \lr (\lambda \identity & + a_j))\tr (\identity - \lr (\lambda \identity + a_j))\hat{\params} \vert \mathcal{F}_{j}]\\
                   &\leq (1 - \lr(2\mu + 2\lambda - \lr(\lambda^2 + 2 (1+\discount) \Feature^{2}_{\mathsf{max}} + (1+\discount)^2\Feature^{4}_{\mathsf{max}})))\norm{\hat{\params}}^{2}_{2}\\
                   &\stackrel{(a)}{\leq}(1 - \lr(2\mu+\lambda))\norm{\hat{\params}}^{2}_{2},\numberthis\label{eq:reg-td-hpbound-abound-1}
               \end{align*}
               where $(a)$ follows from using the using the value of $\lr_\mathsf{max}$ given in \eqref{eq:reg-td-lr}.
               
               \noindent Using Jensen's inequality along with \eqref{eq:reg-td-hpbound-abound-1}, we have
               \begin{align*}
                   \expecun{}\left[\norm{(\identity - \lr(\lambda \identity+a_j))\hat{\params}}_{2} \vert \mathcal{F}_{j}\right] &\leq \left(1 - \lr(\mu+\lambda)\right)^{1/2}\norm{\hat{\params}}_2\\
                   &\stackrel{(a)}{\leq} \left(1-\frac{\lr(\mu+\lambda)}{2}\right)\norm{\hat \params}_{2},
               \end{align*}
                where $(a)$ uses the following fact: $1 - \lr (\mu+\lambda) \geq 0 \implies (1 - \lr (\mu+ \lambda))^{\frac{1}{2}} \leq 1 - \frac{\lr(\mu+\lambda)}{2}$.
            \end{proof}

        \subsubsection{Bias-variance decomposition of the non-asymptotic error} \label{sec:bias-var-decomp-regtd}

        Let 
        \begin{align}
            \hat{z}_{\timestep}^{\mathsf{bias}} &\define \expecun{}\left[\norm{C^{\timestep} \hat{z}_0]}_{2}^{2}\right], \label{eq:z-bias-regtd}\\
            \nonumber&\text{and}\\
            \hat{z}_{\timestep}^{\mathsf{variance}} &\define \sum_{k=0}^{\timestep} \expecun{}\left[\norm{C^{k}\Delta M_{t-k}}_{2}^{2}\right] \label{eq:z-var-regtd}.
        \end{align}
            Now define the centered error rule as $\hat{z}_{\timestep} = \hat{\params}_{\timestep} - \params^{\star}_{\mathsf{reg}}$.
            \noindent Using \cref{eq:req-td-update}, we obtain 
            \begin{align*}
                \hat{z}_{\timestep} &= (\identity-\lr \lambda)\hat{\params}_{\timestep-1} - \params^{\star}_{\mathsf{reg}}  + \lr\bigg(-A\hat{\params}_{\timestep-1} + b + \Delta M_{\timestep}\bigg)\\
                                    &= (\identity-\lr \lambda)\hat{\params}_{\timestep-1} - \params^{\star}_{\mathsf{reg}}  + \lr\bigg(-A\hat{\params}_{\timestep-1} + (A+\lambda\identity) (A+\lambda\identity)^{-1}b + \Delta M_{\timestep}\bigg)\\
                                    &= (\identity - \lr (A + \lambda \identity)) \hat{z}_{\timestep-1} + \lr \Delta M_{\timestep}\\
                                    &\stackrel{(a)}{=} C^{\timestep} \hat{z}_0 + \lr \sum_{k=0}^{\timestep} C^{k}\Delta M_{t-k},\numberthis\label{eq:decomp-regtd}
            \end{align*}
            where $C = (\identity - \lr (A + \lambda \identity))$, and $(a)$ follows from unrolling the update rule.  
            
            \noindent Taking expectation on both sides of \cref{eq:decomp-regtd} we obtain
            \begin{align*}
                \expecun{}\left[ \norm{\hat{z}_{\timestep}}_{2}^{2}\right] &= \expecun{}\left[\norm{C^{\timestep} \hat{z}_0 + \lr \sum_{k=0}^{\timestep} C^{k}\Delta M_{t-k}}_{2}^{2}\right]\\
                                                        &\stackrel{(a)}{\leq}2\hat{z}_{\timestep}^{\mathsf{bias}} + 2\lr^{2}\hat{z}_{\timestep}^{\mathsf{variance}},\numberthis\label{eq:bias-var-bound-regtd}
            \end{align*}
            where $(a)$ is obtained by using the following inequalities i. $\norm{a+b}^{2}_2 \leq 2\norm{a}^{2}_{2} + 2\norm{b}^{2}_2$,  ii. $\norm{\sum_{i=0}^{n} \ip_i} \leq \sum_{i=0}^{n}\norm{\ip_i}$, and $\hat{z}_{\timestep}^{\mathsf{bias}} \& \ \hat{z}_{\timestep}^{\mathsf{variance}}$ are defined in \cref{eq:z-bias-regtd,eq:z-var-regtd}
            
            \noindent Therefore, a bound on $\expecun{}\norm{\hat{z}_{\timestep}}_{2}^{2}$ can obtained by bounding individual terms in \cref{eq:bias-var-bound-regtd}.

        \subsubsection{Bounding $\hat{z}_{\timestep}^{\mathsf{bias}}$}
            \begin{lemma}\label{lemma2-regtd}
                For any step size $\lr \leq \lr_{\mathsf{max}}$, the bias or the initial error of the TD update is upper bounded as
                \begin{align*}
                     \hat{z}_{\timestep}^{\mathsf{bias}} &\leq \exp(-\lr(\mu+\lambda)\timestep)\expecun{}[\norm{\hat{z}_{0}}^{2}_{2}].
                \end{align*}
            \end{lemma}
            \begin{proof}
                \begin{align*}
                     \hat{z}_{\timestep}^{\mathsf{bias}} &= \expecun{}[\norm{C^{\timestep:1} \hat{z}_0]}_{2}^{2}] \\
                     &=\expecun{}\left[(C^{\timestep-1:1}\hat{z}_{0})^{\top}(\identity - \lr (A + \lambda \identity))^{\top}(\identity - \lr (A + \lambda \identity))C^{\timestep-1:1}\hat{z}_{0}\right]\\
                    &\stackrel{(a)}{\leq} \norm{(\identity - \lr (\identity - \lr (A + \lambda \identity)))^{\top}(\identity - \lr (A + \lambda \identity))}_{2}\expecun{}[\norm{C^{t-1:1}\hat{z}_{0}}_{2}^{2}] ,\\
                                                            &\stackrel{(b)}{\leq}(1-\lr(\mu+\lambda))\expecun{}[\norm{C^{t-1:1}\hat{z}_{0}}^{2}_{2}] \numberthis\label{eq:crude-bias-regtd},\\
                                                            &\stackrel{(c)}{=}(1 - \lr(\mu+\lambda))^{t}\expecun{}[\norm{\hat{z}_{0}}^{2}_{2}]\numberthis\label{eq:bias-bound-res1-regtd},\\
                                                            &\leq \exp(-\lr(\mu+\lambda)\timestep)\expecun{}[\norm{\hat{z}_{0}}^{2}_{2}],
                \end{align*}
                where $(a)$ follows Cauchy-Schwarz inequality, $(b)$ follows from \Cref{lem:high-prob-helper-1}, and $(c)$ is the recursive application of the bound in $(b)$.
            \end{proof}

        \subsubsection{Bounding $\hat{z}_{\timestep}^{\mathsf{variance}}$}
            \begin{lemma}\label{lem:var-helper-1-regtd}
            For any random vector $\ip \in \real^{d}$,   
            \begin{align*}
                \sum_{i = 0}^{\timestep}\expecun{}[\norm{C^{i} \ip}^{2}_{2}] &\leq \frac{\expecun{}[\norm{\ip}^{2}_{2}]}{\lr(\mu+\lambda)}.
            \end{align*}
            \end{lemma}
            \begin{proof}
                \begin{align*}
                \expecun{}\left[\sum_{i = 0}^{\timestep}\norm{C^{i} \ip}^{2}_{2}\right] &= \expecun{} \left[\sum_{i = 0}^{\timestep} \left(C^{i} \ip\right)^{\top}
                \left(C^{i} \ip\right)\right]\\
                &= \sum_{i = 0}^{\timestep} \expecun{}\left[ \left(\ip^{\top} (C^{i})^{\top}C^{i} \ip)\right)\right] \\
                &\stackrel{(a)}{\leq} \sum_{i=0}^{\timestep}\norm{(\identity - \lr (\identity - \lr (A + \lambda \identity)))^{\top}(\identity - \lr (\identity - \lr (A + \lambda \identity)))}_{2}\expecun{}\left[\ip^{\top}(C^{i-1})^{\top}C^{i-1}\ip\right]\\
                &\stackrel{(b)}{\leq} \sum_{i=0}^{\timestep}(1 - \lr (\mu+\lambda))\expecun{}\left[\ip^{\top}(C^{i-1})^{\top}C^{i-1}\ip\right]\\
                &\stackrel{(c)}{\leq}\sum_{i=0}^{\timestep}(1-\lr(\mu+\lambda))^{i}\expecun{}\left[\norm{\ip}^{2}_{2}\right]\\
                &\stackrel{(d)}{\leq}\frac{\expecun{}\left[\norm{\ip}^{2}_{2}\right]}{\lr(\mu+\lambda)},
            \end{align*} 
            Where $(a)$ follows from Cauchy-Schwarz inequality, $(b)$ follows from \Cref{lem:high-prob-helper-1} and definition of $C^{k}$, $(c)$ follows from  unrolling the recursion, and $(d)$ follows from the fact that $\sum_{i=1}^{\timestep}(1- \lr (\mu+\lambda))^{i} \leq \frac{1}{ \lr(\mu+\lambda)}$.
                
            \end{proof}
        
            \begin{lemma}\label{lem:var-bound-regtd}
            
             \begin{align*}
                 \hat{z}_{\timestep}^{\mathsf{variance}} &\leq \frac{\sigma^{2}}{\lr(\mu+\lambda)}.
             \end{align*}
            \end{lemma}
            
            \begin{proof}
             
             \begin{align*}
                 \hat{z}_{\timestep}^{\mathsf{variance}} &= 
                 \sum_{i = 1}^{\timestep} \expecun{} \bigg[ \bigg( C^{i} \Delta M_{i-t}\bigg)^{\top} 
                 \bigg( C^{i}\Delta M_{i-t}\bigg)\bigg]\\
                 &\stackrel{(a)}{\leq}\frac{1}{(\mu+\lambda)}\expecun{}[\norm{\Delta M_\timestep}_{2}^{2}| \mathcal{F}_{\timestep-1}]\\
                 &\stackrel{(b)}{\leq}\frac{\sigma^{2}}{ \lr(\mu+\lambda)},
            \end{align*}  
            where $(a)$ follows from \Cref{lem:var-helper-1-regtd}, and $(b)$ follows from \Cref{asm:bddRewards}, and the fact that $\expecun{}[\norm{\Delta M_\timestep}_{2}^{2}| \mathcal{F}_{\timestep-1}] \leq \sigma^2$, where $\sigma^2 = (\Cost_{\mathsf{max}} + (1+\discount)\Feature^{2}_{\mathsf{max}}\norm{\params^{\star}}^{2}_{2})$.
            \end{proof}
            
 	\subsubsection{Proof of Theorem \ref{thm:regtd-expectation_bound}}\label{sec:tailavregtd-bound}
                We will first proceed by getting the bias variance decomposition of the tail-averaged regularised TD update.
                
                \begin{align*}
                 \expecun{}[\norm{\hat{z}_{k+1,N}}_{2}^{2}] &= \frac{1}{N^{2}}\sum_{i,j = k+1}^{k+N}\expecun{}[\hat{z}_i^{\top} \hat{z}_j]\\
                                          &\stackrel{(a)}{\leq} \frac{1}{N^{2}}\bigg(\sum_{i= k+1}^{k+N}\expecun{}[\norm{\hat{z}_i}_{2}^{2}] + 2 \sum_{i=k+1}^{k+N-1}\sum_{j=i+1}^{k+N} \expecun{}[\hat{z}_{i}^{\top}\hat{z}_{j}] \bigg)\numberthis\label{eq:cross-term-decomp-regtd},
                \end{align*}
                where $(a)$ follows from separating out the diagonal and off-diagonal terms and using Cauchy-Schwarz inequality.
                
            \begin{lemma}\label{lem:tav-bias-support-regtd}
            For all $i\ge 1$, we have
            \begin{align*}
                \sum_{i=k+1}^{k+N-1}\sum_{j=i+1}^{k+N} \expecun{}[\hat{z}_{i}^{\top}\hat{z}_{j}]  
                &\leq \frac{2}{\lr (\mu+\lambda)}\sum_{i=k+1}^{k+N}\expecun{}[\norm{\hat{z}_{i}}^{2}_{2}].
            \end{align*}
            \end{lemma}
            \begin{proof}
                \begin{align*}
                   \sum_{i=k+1}^{k+N-1}\sum_{j=i+1}^{k+N} \expecun{} [\hat{z}_{i}^{\top}\hat{z}_{j}]  &=  
                   \sum_{i=k+1}^{k+N-1}\sum_{j=i+1}^{k+N} \expecun{} [\hat{z}_{i}^{\top}(C^{(j-i)}\hat{z}_{i} + \lr\sum_{l=0}^{j-i-1} C^{l}\Delta M_{j - l}))]\\
                   &\stackrel{(a)}{=}\sum_{i=k+1}^{k+N-1}\sum_{j=i+1}^{k+N} \expecun{}  [\hat{z}_{i}^{\top}C^{(j-i)}\hat{z}_{i}] \\
                   &\leq \sum_{i=k+1}^{k+N-1}\sum_{j=i+1}^{k+N} \expecun{}[\norm{\hat{z}_i}_{2}^{2}\norm{C^{(j-i)}}_{2}]\\
                   &\stackrel{(b)}{\leq}\sum_{i=k+1}^{k+N-1}\sum_{j=i+1}^{k+N} (1 - \lr (\mu+\lambda))^{j-i}\expecun{}[\norm{\hat{z}_{i}}^{2}_{2}]\\
                   &\leq \sum_{i=k+1}^{k+N}\expecun{}[\norm{\hat{z}_{i}}^{2}_{2}]\sum_{j=i+1}^{\infty}(1- \lr (\mu+\lambda))^{j-i}\\
                   &\stackrel{(c)}{\leq}\frac{2}{\lr (\mu+\lambda)}\sum_{i=k+1}^{k+N}\expecun{}[\norm{\hat{z}_{i}}^{2}_{2}],
                \end{align*}
                
            where $(a)$ follows from the fact that $\expecun{}[\Delta M_t] = 0$, $(b)$ follows from \Cref{lem:regtd-featurebound}, $(c)$ follows from the summation of the geometric series.
            \end{proof}
        
        For the sake of convenience, we restate the bound in expectation for regularised TD below.
            \begin{theorem}[\textbf{Bound in expectation}] \label{thm:regtd-expectation_bound-2}
                Suppose  \Crefrange{asm:stationary}{asm:bddRewards} hold. Choose a step size $\lr$ satisfying 
                \begin{align*}
                   \lr \leq \lr_{\mathsf{max}} = \frac{\lambda}{\lambda^2 + 2\lambda (1+\discount)\Phi^{2}_{\mathsf{max}} + (1+ \discount)^2 \Phi^4_{\mathsf{max}}}.\numberthis\label{eq:reg-td-lr-2}
                \end{align*}
                Then the expected error of the tail-averaged regularised TD iterate $\hat{\params}_{k+1,N}$ satisfies
            \begin{align*}
               \expecun{}[\norm{\hat{\params}_{k+1,N} - \params^{\star}_{\mathsf{reg}}}^{2}_{2}] &\leq \frac{  10e^{(-k \lr (\mu+\lambda))}}{\lr^2 (\mu+\lambda)^2 N^{2}}\expecun{}\left[\norm{\hat{\params}_{0} - \params^{\star}_{\mathsf{reg}}}^{2}_2\right] + \frac{10\sigma^{2}}{(\mu+\lambda)^2 N},\numberthis\label{eq:regtd-expec-bd-2}
           \end{align*}
               where $N = \timestep - k$, and $\sigma^2 = (\Cost_{\mathsf{max}} + (1+\discount)\Feature^{2}_{\mathsf{max}}\norm{\hat\params^{\star}}_{2}^{2})$.
            \end{theorem}
        
            \begin{proof}
            
                Substituting the result of \Cref{lem:tav-bias-support-regtd} in \cref{eq:cross-term-decomp-regtd} we get
                \begin{align*}
                     \expecun{}[\norm{\hat{z}_{k+1,N}}_{2}^{2}] &\leq \frac{1}{N^{2}}\bigg(\sum_{i = k+1}^{k+N}\expecun{}[\norm{\hat{z}_i}_{2}^{2}] + \frac{4}{\lr
                     (\mu+\lambda)} \sum_{i = k+1}^{k+N}\expecun{}[\norm{\hat{z}_i}_{2}^{2}] \bigg)\\
                                                         &= \frac{1}{N^{2}}\bigg(1+\frac{4}{\lr (\mu+\lambda)}\bigg) \sum_{i = k+1}^{k+N}\expecun{}[\norm{\hat{z}_i}_{2}^{2}]\\
                                                         &\stackrel{(a)}{\leq} \underbrace{\frac{2}{N^{2}}\bigg(1+\frac{4}{\lr (\mu+\lambda)}\bigg) \sum_{i = k+1}^{k+N}   \hat{z}_{i}^{\mathsf{bias}}}_{\hat{z}_{k+1, N}^{\mathsf{bias}}} + \underbrace{\frac{2}{N^{2}}\bigg(1+\frac{4}{\lr (\mu+\lambda)}\bigg)\lr^{2}  \sum_{i = k+1}^{k+N}   \hat{z}_{i}^{\mathsf{variance}}}_{\hat{z}_{k+1,N}^{\mathsf{variance}}}\numberthis\label{eq:tail-av-bias-var-2-regtd}.
                \end{align*}
                where $(a)$ follows from \cref{eq:bias-var-bound-regtd}, and $N = \timestep - k$.
                
                \noindent $\hat{z}_{k+1,N}^{\mathsf{bias}}$in \cref{eq:tail-av-bias-var-2-regtd}  is bounded as follows
                \begin{align*}
                            \hat{z}_{k+1,N}^{\mathsf{bias}} &\leq \frac{2}{N^{2}}\left(1 + \frac{4}{\lr (\mu+\lambda)}\right) \sum_{i = k+1}^{\infty}   \hat{z}_{i}^{\mathsf{bias}} \\
                            &\stackrel{(a)}{\leq} \frac{2}{N^{2}}\left(1 + \frac{4}{\lr (\mu+\lambda)}\right) \sum_{i = k+1}^{\infty} \left(1 - \lr (\mu+\lambda)\right)^{i}\expecun{}[\norm{\hat{z}_{0}}^{2}_{2}]  \\
                            &\stackrel{(b)}{=}\frac{2}{\lr (\mu+\lambda) N^{2}}(1-\lr(\mu+\lambda))^{k+1}\left(1 + \frac{4}{\lr (\mu+\lambda)}\right)\expecun{}[\norm{\hat{z}_{0}}^{2}_{2}],
                        \end{align*}
                   where $(a)$ follows from \cref{eq:bias-bound-res1-regtd} in proof of \Cref{lemma2-regtd}, and $(b)$ follows from the summation of the geometric series.
                   
                \noindent $\hat{z}_{k+1,N}^{\mathsf{variance}}$ in \cref{eq:tail-av-bias-var-2-regtd} is bounded as follows
                \begin{align*}
                        \hat{z}_{k+1,N}^{\mathsf{variance}}
                        &\stackrel{(a)}{\leq}\frac{2\lr^2}{N^{2}}\left(1+\frac{4}{\lr (\mu+\lambda)}\right)\sum_{i=k+1}^{k+N}\frac{\sigma^2}{\lr(\mu+\lambda)}\\
                        &\leq \frac{2\lr}{N^{2}}\left(1+\frac{4}{\lr (\mu+\lambda)}\right)\sum_{i=0}^{N}\frac{\sigma^2}{(\mu+\lambda)}\\
                        &= \bigg(1+\frac{4}{\lr (\mu+\lambda)}\bigg)\frac{2\lr\sigma^2}{(\mu+\lambda) N},
                    \end{align*}
                    where  $(a)$ follows from \Cref{lem:var-bound-regtd}.
                
              \noindent Finally substituting the bounds on $\hat{z}_{k+1,N}^{\mathsf{bias}}$ and $\hat{z}_{k+1,N}^{\mathsf{variance}}$ in \eqref{eq:tail-av-bias-var-2-regtd}, we get
               \begin{align*}
                \expecun{}\left[\norm{\hat{z}_{k+1,N}}^{2}_{2}\right] &\leq\bigg(1+ \frac{4}{\lr(\mu+\lambda)}\bigg)\bigg(\frac{2}{\lr (\mu+\lambda) N^{2}}(1-\lr(\mu+\lambda))^{k+1}\expecun{}\left[\norm{\hat{z}_{0}}^{2}_{2}\right] + \frac{2\lr\sigma^2}{(\mu+\lambda) N}\bigg),\\
                &\stackrel{(a)}{\leq} \bigg(1+ \frac{4}{\lr(\mu+\lambda)}\bigg)\bigg(\frac{2 \exp(-k \lr (\mu+\lambda))}{\lr (\mu+\lambda) N^{2}}\expecun{}[\norm{\hat{z}_{0}}^{2}_{2}] + \frac{2\lr\sigma^2}{(\mu+\lambda) N}\bigg)\\
                &\stackrel{(b)}{\leq} \frac{10 \exp(-k \lr (\mu+\lambda))}{\lr^2 (\mu+\lambda)^2 N^{2}}\expecun{}\left[\norm{\hat{z}_{0}}^{2}_{2}\right] + \frac{10\sigma^2}{(\mu+\lambda)^2 N},
                \end{align*}
            where $(a)$ is because $(1+x)^{y} = \exp(y\log(1+x)) \leq \exp(xy)$, and $(b)$ is because $1+ \frac{4}{\lr (\mu+\lambda)} \leq \frac{5}{\lr (\mu+\lambda)}$
        \end{proof}
        
        \begin{corollary}\label{cor:reg-td-2}
        Under conditions of Theorem \ref{thm:regtd-expectation_bound}, we have
            \begin{align*}
                 \expecun{}\left[\norm{\hat\params_{k+1,N} - \params^{\star}}_{2}^{2}\right] &\leq  \frac{  20e^{(-k \lr (\mu+\lambda))}}{\lr^2 (\mu+\lambda)^2 N^{2}}\expecun{}\left[\norm{\hat{\params}_{0} - \params^{\star}_{\mathsf{reg}}}^{2}_2\right] + \frac{20\sigma^{2}}{(\mu+\lambda)^2 N} + \frac{2 \lambda^2 \Feature^{2}_\mathsf{max}R_\mathsf{max}^{2}}{\sigma_{\mathsf{min}}(A)^2(\lambda +\mu)^2},
            \end{align*}
        where $\sigma_{\text{min}}(A)$ as $A$'s minimum singular value.
        \end{corollary}
        \begin{proof}\label{proof:cor1}
            \begin{align*}
                 \expecun{}\left[\norm{\hat\params_{k+1,N} - \params^{\star}}_{2}^{2}\right] 
                    &\stackrel{(a)}{\leq}   \underbrace{2\expecun{}\left[\norm{\params^{\star}_{\mathsf{reg}} - \params^{\star}}_{2}^{2}\right]}_{\text{Term 1}} + \underbrace{2\expecun{}\left[\norm{\hat\params_{k+1,N} - \params^{\star}_{\mathsf{reg}}}_{2}^{2}\right])}_{\text{Term 2}},\numberthis\label{eq:regtd-drift-crude}
            \end{align*}
            where $(a)$ is because $\Vert a + b \Vert^{2}_{2} \leq 2 \Vert a \Vert_{2}^{2} +  2 \Vert b \Vert_{2}^{2} $.
            
            \noindent Bounding Term 1
            \begin{align*}
                \expecun{}\left[\norm{\params^{\star} - \params^{\star}_{\mathsf{reg}}}_{2}^{2}\right] &= \norm{A^{-1}b - (A + \lambda \identity)^{-1}b}_{2}^{2}\\
                &\stackrel{(a)}{\leq} \norm{A^{-1} - (A + \lambda \identity)^{-1}}_{2}^{2}\norm{b}_{2}^{2}\\
                &\stackrel{}{=} \norm{A^{-1} (A + \lambda \identity - A) (A + \lambda \identity)^{-1}}_{2}^{2}\norm{b}_{2}^{2}\\
                &\stackrel{(a)}{\leq} \norm{A^{-1}}_{2}^{2} \lambda^2 \norm{(A + \lambda \identity)^{-1}}_{2}^{2}\norm{b}_{2}^{2}\\
                &\stackrel{(b)}{\leq}\frac{\lambda^2\Feature_{\mathsf{max}}^{2}}{\sigma_{\mathsf{min}}(A)^2(\lambda +\mu)^2},\numberthis\label{eq:regtd-drift}
            \end{align*}
             where $(a)$  follows from Cauchy-Schwarz inequality, and $(b)$ follows from the fact that  $\norm{A^{-1}} = 1/\sigma_{\text{min}}(A)$, with $\sigma_{\text{min}}(A)$ as $A$'s minimum singular value.
            
            
        \end{proof}

        \subsection{High probability bound for Tail-averaged TD with regularisation }\label{sec:regtd-highprob-proof}
        
            \begin{proposition}\label{prop:hprob-regtd}
                Under assumptions \ref{asm:bddFeatures}, \ref{asm:bddRewards}, for all $\epsilon \geq 0$, and $\timestep \geq 1$,
                \begin{align*}
                    \prob\left(\norm{\hat{z}_{k+1,N}}_{2} - \expecun{}\left[\norm{\hat{z}_{k+1,N}}_{2} > \epsilon\right] \right) \leq \exp\left(-\frac{\epsilon^{2}}{ (\Cost_{\mathsf{max}} + (1+ \discount) H \Feature^{2}_{\mathsf{max}})^{2} \sum_{i = k+1}^{k+N} L^{2}_{i}} \right),
                \end{align*}
            where $L_i = \frac{\lr}{N}\sum_{j=i+1}^{i+N}\left(1 - \frac{\lr (\mu+\lambda)}{2}\right)^{j-i+1}$.
            \end{proposition}
                \noindent For this proof we can use Step 1 and Step 3 of \Cref{prop:hprob}. However, as the update rule is differs from the usual TD update, we will derive the Lipschitz constant (Step 2 in proof of \Cref{prop:hprob}) for the regularised TD algorithm separately. Towards that end, 
                
                 \noindent We need to prove that that functions $g_{i}$ are Lipschitz continuous in $f_i$ at time $i$ with new constant $L_{i}$.

                 Towards that end, first, define $\hat{\Theta}^{i}_{\timestep,k+1}(\params)$ to be the value of the regularised-tail-averaged iterate at time $\timestep$ that evolves according to \cref{eq:td-update} beginning from $\params$ at time $i$, and next, define $\bar{\hat{\Theta}}^{i}_{k+1,N}(\bar{\hat\params}, \hat{\params})$ as follows:
                
                \begin{align*}
                    \bar{\hat{\Theta}}^{i}_{k+1,N}(\bar{\hat\params}, \hat{\params}) &= \frac{(i-k)\bar{\hat{\params}}}{N} + \frac{1}{N}\sum_{j = i+1}^{i+N}\hat{\Theta}^{i}_{j}(\hat{\params}),
                \end{align*}
                where $\hat{\params}$ is the value of the tail-averaged regularised TD iterate at time $i$. 

                 \noindent Let $f$ and $f'$ denote two different values of random innovations at time $i$. Then we know that, $\hat{\params} = \hat{\params}_{i-1} + \lr f$, and $\hat{\params}' = \hat{\params}_{i-1} + \lr f'$ are the parameter values corresponding to each $f$. Therefore,

                \begin{align*}
                    \expecun{}\left[\norm{\bar{\hat{\Theta}}^{i}_{i+1,N}(\bar{\hat{\params}}, \hat{\params}) -  \bar{\hat{\Theta}}^{i}_{i+1,N}(\bar{\hat{\params}}, \hat{\params}') }_{2}\right] &=
                    \expecun{}\left[\frac{1}{N}\sum_{j=i+1}^{i+N}\norm{ \hat{\Theta}^{i}_{j}(\hat{\params}) - \hat{\Theta}^{i}_{j} (\hat{\params}')}_{2}\right].\numberthis\label{eq:interm-reg-td}
                \end{align*}
                
                \noindent We will now bound the term $\hat{\Theta}^{i}_{j}(\hat{\params}) - \hat{\Theta}^{i}_{j}(\hat{\params}')$ inside the summation of \eqref{eq:interm-reg-td}.

                Note that as the projection $\Gamma$ is non-expansive, we have the following
                \begin{align*}
                    \expecun{}\bigg[\norm{\hat{\Theta}_{j}^{i}(\hat{\params}) - \hat{\Theta}_{j}^{i}(\hat{\params}')}_{2}\bigg \vert\mathcal{F}_{j-1}\bigg]
                    &\leq  \ \expecun{}\bigg[ \norm{ \big(\hat{\Theta}_{j-1}^{i}(\hat{\params}) - \hat{\Theta}_{j-1}^{i}(\hat{\params}')\big) - 
                    \lr [f_{i}(\hat{\Theta}_{j-1}^{i}(\hat{\params})) - f_{i}(\hat{\Theta}_{j-1}^{i}(\hat{\params}')])}_{2} \bigg \vert \mathcal{F}_{j-1}\bigg].
                \end{align*}
                  
                \noindent Expanding on $f_i$ and using$a_{j} \define [\feature(\state_{j})\feature(\state_{j}){\tr} - \discount\feature(\state_{j})\feature(\state'_{j}){\tr}]$ , we have
                
                \begin{align*}
                    \nonumber&\hat{\Theta}_{j-1}^{i}(\hat{\params}) - \hat{\Theta}_{j}^{i}(\hat{\params}') - \lr [f_{i}(\hat{\Theta}_{j-1}^{i}(\hat{\params})) - f_{i}(\hat{\Theta}_{j-1}^{i}(\hat{\params}')]
                     = \\
                     &(1- \lr\lambda \identity )\big(\hat{\Theta}_{j-1}^{i}(\hat{\params}) - \hat{\Theta}^{i}_{j-1}(\hat{\params}')\big) - \lr [\feature(\state_{j})\feature(\state_{j}){\tr} - \discount\feature(\state_{j})\feature(\state'_{j}){\tr} ] [(\hat{\Theta}^{j}_{i-1}(\hat{\params})) - (\hat{\Theta}^{j}_{i-1}(\hat{\params}')]\\
                     &= [\identity - \lr (a_{j} + \lambda \identity)](\hat{\Theta}_{j-1}^{i}(\hat{\params}) - \hat{\Theta}_{j-1}^{i}(\hat{\params}')). \numberthis\label{eq:step-2-interm-regtd}
                \end{align*}
            
            \noindent Using the tower property of conditional expectations, it follows that:
                \begin{align*}
                     \expecun{}\left[\norm{\hat{\Theta}^{i}_{j}(\hat{\params}) - \hat{\Theta}^{i}_{j}(\hat{\params}')}_{2} \right] &= \expecun{}\bigg[\expecun{}\left[\norm{\hat{\Theta}^{i}_{j}(\hat{\params}) - \hat{\Theta}^{i}_{j}(\hat{\params}')}_{2} \bigg\vert \mathcal{F}_{j-1}\right]\bigg]\\
                     &= \expecun{}\bigg[\expecun{}\left[\norm{(\identity - (\lambda \identity + a_j))\hat{\Theta}^{i}_{j-1}(\hat{\params}) - \hat{\Theta}^{i}_{j-1}(\hat{\params}')}_{2}\bigg \vert \mathcal{F}_{j-1}\right]\bigg]\\
                    &\stackrel{(a)}{\leq} \left(1-\frac{\lr(\mu+\lambda)}{2}\right) \expecun{}\left[\norm{\hat{\Theta}^{i}_{j-1}(\hat{\params}) - \hat{\Theta}^{i}_{j-1}(\hat{\params}')}_{2}\bigg \vert \mathcal{F}_{j-1}\right]\\
                    &\stackrel{(b)}{\leq} \left(1 - \frac{\lr(\mu+\lambda)}{2}\right)^{j-i+1} \norm{\hat{\params} - \hat{\params}'}_{2},
                \end{align*}
                where $(a)$ follows from \Cref{lem:regtd-highprob-helper1}, and $(b)$ follows from repeated application of argument that helps us arrive at $(a)$. 
                
            Now using Jensen's inequality we get,
                
            \begin{align*}
                \bigg| \expecun{}\left[\norm{\hat{\params}_j - \params^{\star}_{\mathsf{reg}}}_{2}|\hat{\params}_{j} = \hat{\params}\right] - \expecun{}\left[\norm{\hat{\params}_j - \params^{\star}_{\mathsf{reg}}}_{2}|\hat{\params}_{j} = \hat{\params}'\right]\bigg| &\leq \expecun{}\left[\norm{\hat{\Theta}^{i}_{j}(\hat{\params}) - \hat{\Theta}^{i}_{j}(\hat{\params}')}_{2}\right]
                \leq \lr\left(1 - \frac{\lr (\mu+\lambda)}{2}\right)^{j-i+1} \norm{f - f'}_{2}\numberthis\label{eq:lip-comp-regtd}.\\
            \end{align*}

            \noindent Substituting \eqref{eq:lip-comp-regtd} in \eqref{eq:interm-reg-td}, and using \Cref{lem:tav-bias-support}, we get the following
            
            \begin{align*}
                \expecun{}\left[\norm{\bar{\hat{\Theta}}^{i}_{k+1}(\bar{\hat{\params}}_{i-1}, \hat{\params}) -  \bar{\hat{\Theta}}^{i}_{k+1}(\bar{\hat{\params}}_{i-1}, \hat{\params}') }_{2}\right] 
                &\leq \frac{\lr}{N}\sum_{j=i+1}^{i+N}\left(1 - \frac{\lr (\mu+\lambda)}{2}\right)^{j-i+1} \norm{f - f'}_{2}. \numberthis\label{eq:lip-up-regtd}
            \end{align*}
            
            \noindent From \eqref{eq:lip-up-regtd} it is clear that $g_i$ is $L_i$-Lipschtiz in $f_i$ at time $i$, which implies that $D_i$ is Lipschitz with Lipschitz constant $L_i = \frac{\lr}{N}\sum_{j=i+1}^{i+N}\left(1 - \frac{\lr (\mu+\lambda)}{2}\right)^{j-i+1}$.

            \subsubsection{Bounding the Lipschitz constant}
            
            \begin{lemma}\label{lem:lipschitz-regtd}
                For the tail-averaged TD, the Lipschitz constant $L_{i}$ in \Cref{prop:hprob} is upper bounded as follows:
                \begin{align*}
                    \sum_{i=k+1}^{k+N} L_{i}^{2} &\leq \frac{4}{(\mu + \lambda)^{2}N}.
                \end{align*}
            \end{lemma}
            \begin{proof}
                 \begin{align*}
                     \sum_{i=k+1}^{k+N} L_{i}^{2} &=\frac{\lr^2}{N^2}\sum_{i=k+1}^{k+N} \left(\sum_{j=i+1}^{i+N}\left(1 - \frac{\lr (\mu+\lambda)}{2}\right)^{j-i+1}\right)^{2}\\
                    &\leq \frac{\lr^2}{N^2}\sum_{i=k+1}^{k+N} \left(\sum_{j=i+1}^{\infty}\left(1 - \frac{\lr (\mu+\lambda)}{2}\right)^{j-i+1}\right)^{2}\\
                    &\stackrel{(a)}{=}\frac{\lr^2}{N^2}\sum_{i=k+1}^{k+N}\frac{4}{\lr^2(\mu + \lambda)^{2}} \\
                    &\leq \frac{\lr^2}{N^2}\sum_{i=k+1}^{\infty}\frac{4}{\lr^2( \mu + \lambda)^{2}}\\
                    &= \frac{4}{(\mu + \lambda)^{2}N},
                \end{align*}
            where (a) is due to summing the geometric series.
            \end{proof}

            \subsubsection{Proof of Theorem  \ref{thm:regtd-high-prob}}
            For the sake of convenience, we restate the high probability bound for regularised TD below.
            \begin{theorem}[\textbf{High-probability bound}]\label{thm:regtd-high-prob-2}
            	Assume \ref{asm:stationary} to \ref{asm:bddRewards}, and \ref{asm:projection}. Choose the step size such that $\lr \leq \lr_{\mathsf{max}}$, where $\lr_{\mathsf{max}}$ is defined in \eqref{eq:reg-td-lr}. Then, for any  $\delta\in (0,1]$, we have the following bound for the projected tail-averaged regularised TD iterate $\hat{\params}_{k+1,N}$:
            \begin{align*}
                \prob\left(\norm{\hat{\params}_{k+1,N} - \params^{\star}_{\mathsf{reg}}}_{2} 
                \leq 
                \frac{2\sigma}{(\mu+\lambda)\sqrt{N}}\sqrt{\log\left(\frac{1}{\delta}\right)} +  \frac{4e^{(-k \lr (\mu + \lambda))}}{\lr(\mu+\lambda) N}\expecun{}\left[\norm{\hat{\params}_{0} - \params^{\star}_{\mathsf{reg}}}_{2}\right] +  \frac{4\sigma}{(\mu + \lambda)\sqrt{N}}\right) &\geq 1 - \delta,
            \end{align*}
            where $N,\sigma,\mu, \hat{\params}_0,\params^{\star}_{\mathsf{reg}}$ are as specified in Theorem \ref{thm:regtd-expectation_bound}.
            \end{theorem}
            \begin{proof}\label{proof:thm:regtd-high-prob}

                \begin{align*}
                    \prob(\norm{\hat{z}_{k+1,N}}_{2}^{2} - \expecun{}[\norm{\hat{z}_{k+1,N}}_{2}^{2}] > \epsilon ) &\stackrel{(a)}{\leq} \exp\bigg(-\frac{\epsilon^{2}}{ (\Cost_{\mathsf{max}} + (1+ \discount) H \Feature^{2}_{\mathsf{max}})^{2} \sum_{i = k+1}^{k+N} L^{2}_{i}} \bigg),\\
                    &\stackrel{(b)}{\leq} \exp\left(-\frac{N(\mu + \lambda)^{2}\epsilon^2}{4\sigma^2}\right)\numberthis\label{eq:crude_hprob-redtd},
                \end{align*}
            where $(a)$ follows from \Cref{prop:hprob-regtd}, and $(b)$ follows from \Cref{lem:lipschitz-regtd}.
            
                \noindent The main claim follows by converting the bound above to a high-confidence form. In particular, 
                    \begin{align*}
                        \exp\left(-\frac{N(\mu + \lambda)^{2}\epsilon^{2}}{4\sigma^2}\right) &= \delta, \textrm{ leads to }\quad
                       \epsilon= \frac{2\sigma}{(\mu + \lambda)}\sqrt{\frac{\log\left(\frac{1}{\delta}\right)}{N}}\numberthis\label{eq:epsilon-regtd}.
                    \end{align*}
                The final bound follows by substituting the value for $\epsilon$ from \eqref{eq:epsilon-regtd} in \eqref{eq:crude_hprob-redtd}, and using the result from \Cref{thm:regtd-expectation_bound}.
            \end{proof}

\section{Bounds for Under Mixing Assumptions}
\label{sec:mixing}
Instead of Assumption~\ref{asm:iidNoise}, we now consider the case when $(s_t)_{t\in \mathbb{N}}$ are drawn from a single stationary trajectory of the Markov chain with policy $\policy$. We assume exponential ergodicity for the Markov chain, which holds when any finite Markov chain is irreducible. Let $\stationarydist$ be the corresponding stationary distribution.

\begin{assumption}\label{asm:ergodicity}
$s_1 \sim \rho$ and there exist constants $C$ and $\tmix$ such that for every $t,\tau \in \mathbb{N}$ $$D(\tau) := \sup_{s \in \statespace}\tv(s_{t+\tau}|s_{t} = s,\rho) \leq C \exp(-\tfrac{\tau}{\tmix}),$$
where $\tv$ denotes the total variation distance between probability measures.
\end{assumption}

This is a standard assumption in the literature \citep{Bhandari0S18,srikant19a}. We now adapt Lemma 3 from \citep{nagaraj2020least} to our present setting. 

\begin{lemma}[Adaptation of Lemma 3 in \citep{nagaraj2020least}]\label{lem:mix_couple}
For any $K\in \mathbb{N}$, define the random variable $$S_{K,n} := ((s_1,s_2),(s_{K+1},s_{K+2}),(s_{2K+1},s_{2K+2}),\dots,(s_{nK+1},s_{nK+2})).$$
Let $P$ denote the transition kernel for the Markov chain under policy $\pi$. By $\stationarydist^{(2)}$ denote the joint distribution of $(s_1,s_2)$. Under\Cref{asm:ergodicity}, we have
\[\tv(S_{K,n}, (\stationarydist^{(2)})^{\otimes n}) \leq n D(K-1) \leq nC\exp(-\tfrac{K-1}{\tmix}).\]

Now, let $R_{K,n} = (r_{1},r_{K+1},\dots,r_{nK+1})$ be the random rewards corresponding to $S_{K,n}$ and consider i.i.d random variables $\tilde{S}_{K,n} = ((\tilde{s}_1,\tilde{s}_2),(\tilde{s}_{K+1},\tilde{s}_{K+2}),(\tilde{s}_{2K+1},\tilde{s}_{2K+2}),\dots,(\tilde{s}_{nK+1},\tilde{s}_{nK+2})) \sim (\stationarydist^{(2)})^{\otimes n} $ along with the corresponding rewards $\tilde{R}_{K,n}$. We can define these random variables on a common probability space such that
$$\mathbb{P}((S_{K,n},R_{K,n}) \neq (\tilde{S}_{K,n},\tilde{R}_{K,n})) \leq n D(K-1) \leq nC\exp(-\tfrac{K-1}{\tmix}).$$

\end{lemma}
We will use the mixing technique used in SGD-DD in \citep{nagaraj2020least}. Here, when we obtain samples $(s_t,r_t,s_{t+1})$ from a trajectory instead of from the i.i.d distribution as considered before. we modify the Algorithm~\ref{alg:ciac-a} in the following ways to account for mixing. We fix $K \in \mathbb{N}$.
\paragraph{Modification 1:}

Run Algorithm~\ref{alg:ciac-a} with data $S_{K,n},R_{K,n}$ - i.e, we input $(s_{tK+1},r_{tK+1},s_{tK+2})$ at step $t$. 

\paragraph{Modification 2:}

Run Algorithm~\ref{alg:ciac-a} with data $\tilde{S}_{K,n},\tilde{R}_{K,n}$.

Note that the Modification 2 is exactly same as running the algorithm under Assumption~\ref{asm:iidNoise} for $n$ steps and therefore the results of ~\Cref{thm:high-prob-bound} apply to this case if we replace $N$ with $n$. By the results in Lemma~\ref{lem:mix_couple}, we conclude that the trajectories $(\theta_t)$ generated by modification 1 and $(\tilde{\theta}_t)$ generated by modification 2 can be coupled such that $$\mathbb{P}\left[(\theta_t)_{t=1}^{n+1} \neq (\tilde{\theta}_t)_{t=1}^{n+1}\right] \leq n D(K-1).$$

This is based on the fact that whenever the algorithm is fed with the same input, we obtain the same output. Setting $K = \tmix \log (\tfrac{Cn}{\delta})$, we conclude that under Assumption~\ref{asm:ergodicity}, we have:

$$\mathbb{P}\left[(\theta_t)_{t=1}^{n+1} \neq (\tilde{\theta}_t)_{t=1}^{n+1}\right] \leq \delta.$$

Therefore, we conclude the bounds in Remark~\ref{rem:markov}.
\section{Conclusions}\label{sec:conclusion}
We presented a finite time analysis of tail-averaged TD algorithm with/without regularisation. Our bounds are easy to interpret, and improve the previously known results. To the best of our knowledge, this is the first result that establishes a $O\left(\frac{1}{t}(\right)$ convergence rate for a TD algorithm with a universal step size. Finally, we also analyse TD with regularisation and show how it can be useful in certain problem instances with ill-conditioned features.
\newpage
\bibliography{references}

\begin{thebibliography}{25}
\providecommand{\natexlab}[1]{#1}
\providecommand{\url}[1]{\texttt{#1}}
\expandafter\ifx\csname urlstyle\endcsname\relax
  \providecommand{\doi}[1]{doi: #1}\else
  \providecommand{\doi}{doi: \begingroup \urlstyle{rm}\Url}\fi

\bibitem[Agarwal et~al.(2022)Agarwal, Chaudhuri, Jain, Nagaraj, and
  Netrapalli]{agarwal}
N.~Agarwal, S.~Chaudhuri, P.~Jain, D.~M. Nagaraj, and P.~Netrapalli.
\newblock {Online Target Q-learning with Reverse Experience Replay: Efficiently
  finding the Optimal Policy for Linear MDPs}.
\newblock In \emph{The Tenth International Conference on Learning
  Representations, {ICLR}}, 2022.

\bibitem[Bhandari et~al.(2018)Bhandari, Russo, and Singal]{Bhandari0S18}
J.~Bhandari, D.~Russo, and R.~Singal.
\newblock A finite time analysis of temporal difference learning with linear
  function approximation.
\newblock In \emph{Conference On Learning Theory (COLT)}, volume~75 of
  \emph{Proceedings of Machine Learning Research}, pages 1691--1692. {PMLR},
  2018.

\bibitem[Chen et~al.(2020)Chen, Devraj, Busic, and Meyn]{devraj}
S.~Chen, A.~Devraj, A.~Busic, and S.~Meyn.
\newblock {Explicit Mean-Square Error Bounds for Monte-Carlo and Linear
  Stochastic Approximation}.
\newblock In \emph{International Conference on Artificial Intelligence and
  Statistics (AISTATS)}, volume 108 of \emph{Proceedings of Machine Learning
  Research}, pages 4173--4183. PMLR, 2020.

\bibitem[Chen et~al.(2022{\natexlab{a}})Chen, Clarke, and
  Maguluri]{chen2022target}
Z.~Chen, J.~P. Clarke, and S.~T. Maguluri.
\newblock {Target Network and Truncation Overcome The Deadly triad in $ Q
  $-Learning}.
\newblock \emph{arXiv preprint arXiv:2203.02628}, 2022{\natexlab{a}}.

\bibitem[Chen et~al.(2022{\natexlab{b}})Chen, Zhang, Doan, Clarke, and
  Maguluri]{chen2022finite}
Z.~Chen, S.~Zhang, T.~T. Doan, J.~P. Clarke, and S.~T. Maguluri.
\newblock {Finite-sample analysis of nonlinear stochastic approximation with
  applications in reinforcement learning}.
\newblock In \emph{Automatica}, volume 146, 2022{\natexlab{b}}.

\bibitem[Dalal et~al.(2018)Dalal, Sz{\"{o}}r{\'{e}}nyi, Thoppe, and
  Mannor]{DalalSTM18}
G.~Dalal, B.~Sz{\"{o}}r{\'{e}}nyi, G.~Thoppe, and S.~Mannor.
\newblock Finite sample analyses for {TD(0)} with function approximation.
\newblock In \emph{{AAAI} Conference on Artificial Intelligence (AAAI)}, pages
  6144--6160, 2018.

\bibitem[Durmus et~al.(2021)Durmus, Moulines, Naumov, Samsonov, and
  Wai]{Durmus2021OnTS}
A.~Durmus, {\'E}.~Moulines, A.~Naumov, S.~Samsonov, and H.-T. Wai.
\newblock On the stability of random matrix product with markovian noise:
  Application to linear stochastic approximation and td learning.
\newblock In \emph{COLT}, 2021.

\bibitem[Fathi and Frikha(2013)]{fathi2013transport}
M.~Fathi and N.~Frikha.
\newblock Transport-entropy inequalities and deviation estimates for stochastic
  approximation schemes.
\newblock \emph{Electronic Journal of Probability}, 18:\penalty0 1--36, 2013.

\bibitem[Hu and Syed(2019)]{Mkvjump}
B.~Hu and U.~A. Syed.
\newblock Characterizing the exact behaviors of temporal difference learning
  algorithms using markov jump linear system theory.
\newblock In \emph{NeurIPS}, 2019.

\bibitem[Jaakkola et~al.(1994)Jaakkola, Jordan, and Singh]{JaakkolaJS94}
T.~S. Jaakkola, M.~I. Jordan, and S.~P. Singh.
\newblock On the convergence of stochastic iterative dynamic programming
  algorithms.
\newblock \emph{Neural Comput.}, 6\penalty0 (6):\penalty0 1185--1201, 1994.

\bibitem[Jain et~al.(2018)Jain, Kakade, Kidambi, Netrapalli, and
  Sidford]{jain2018parallelizing}
P.~Jain, S.~Kakade, R.~Kidambi, P.~Netrapalli, and A.~Sidford.
\newblock Parallelizing stochastic gradient descent for least squares
  regression: mini-batching, averaging, and model misspecification.
\newblock \emph{Journal of Machine Learning Research}, 18, 2018.

\bibitem[Lakshminarayanan and Szepesvari(2018)]{lakshminarayanan18a}
C.~Lakshminarayanan and C.~Szepesvari.
\newblock Linear stochastic approximation: How far does constant step-size and
  iterate averaging go?
\newblock In \emph{International Conference on Artificial Intelligence and
  Statistics (AISTATS)}, volume~84 of \emph{Proceedings of Machine Learning
  Research}, pages 1347--1355. PMLR, Apr 2018.

\bibitem[Nagaraj et~al.(2020)Nagaraj, Wu, Bresler, Jain, and
  Netrapalli]{nagaraj2020least}
D.~Nagaraj, X.~Wu, G.~Bresler, P.~Jain, and P.~Netrapalli.
\newblock Least squares regression with markovian data: Fundamental limits and
  algorithms.
\newblock \emph{Advances in neural information processing systems},
  33:\penalty0 16666--16676, 2020.

\bibitem[Patil et~al.(2023)Patil, L.A., Nagaraj, and
  Precup]{pmlr-v206-patil23a}
G.~Patil, P.~L.A., D.~Nagaraj, and D.~Precup.
\newblock Finite time analysis of temporal difference learning with linear
  function approximation: Tail averaging and regularisation.
\newblock In F.~Ruiz, J.~Dy, and J.-W. van~de Meent, editors, \emph{Proceedings
  of The 26th International Conference on Artificial Intelligence and
  Statistics}, volume 206 of \emph{Proceedings of Machine Learning Research},
  pages 5438--5448. PMLR, 25--27 Apr 2023.
\newblock URL \url{https://proceedings.mlr.press/v206/patil23a.html}.

\bibitem[Pineda(1997)]{Pineda97}
F.~J. Pineda.
\newblock Mean-field theory for batched-td(l).
\newblock \emph{Neural Comput.}, 9\penalty0 (7):\penalty0 1403--1419, 1997.

\bibitem[Polyak and Juditsky(1992)]{polyak1992acceleration}
B.~T. Polyak and A.~B. Juditsky.
\newblock Acceleration of stochastic approximation by averaging.
\newblock \emph{SIAM Journal on Control and Optimization}, 30\penalty0
  (4):\penalty0 838--855, 1992.

\bibitem[Prashanth et~al.(2021)Prashanth, Korda, and Munos]{lstd-prashanth}
L.~A. Prashanth, N.~Korda, and R.~Munos.
\newblock Concentration bounds for temporal difference learning with linear
  function approximation: the case of batch data and uniform sampling.
\newblock \emph{Mach. Learn.}, 110\penalty0 (3):\penalty0 559--618, 2021.

\bibitem[Ruppert(1991)]{ruppert1991stochastic}
D.~Ruppert.
\newblock Stochastic approximation.
\newblock \emph{Handbook of Sequential Analysis}, pages 503--529, 1991.

\bibitem[Schapire and Warmuth(2004)]{Schapire2004OnTW}
R.~Schapire and M.~K. Warmuth.
\newblock On the worst-case analysis of temporal-difference learning
  algorithms.
\newblock \emph{Machine Learning}, 22:\penalty0 95--121, 2004.

\bibitem[Srikant and Ying(2019)]{srikant19a}
R.~Srikant and L.~Ying.
\newblock {Finite-Time Error Bounds For Linear Stochastic Approximation and TD
  Learning}.
\newblock In \emph{Conference on Learning Theory (COLT)}, volume~99 of
  \emph{Proceedings of Machine Learning Research}, pages 2803--2830. PMLR,
  2019.

\bibitem[Sutton(1988)]{Sutton88}
R.~S. Sutton.
\newblock Learning to predict by the methods of temporal differences.
\newblock \emph{Mach. Learn.}, 3:\penalty0 9--44, 1988.

\bibitem[Sutton and Barto(2018)]{Sutton+Barto:1998}
R.~S. Sutton and A.~G. Barto.
\newblock \emph{Reinforcement Learning: An Introduction}.
\newblock MIT Press, 2018.
\newblock ISBN 0-262-19398-1.

\bibitem[Tsitsiklis and Van~Roy(1997)]{bvrtd}
J.~N. Tsitsiklis and B.~Van~Roy.
\newblock An analysis of temporal-difference learning with function
  approximation.
\newblock \emph{{IEEE} Transactions on Automatic Control}, 42\penalty0
  (5):\penalty0 674--690, 1997.

\bibitem[Wang and Giannakis(2020)]{WangG20a}
G.~Wang and G.~B. Giannakis.
\newblock {Finite-Time Error Bounds for Biased Stochastic Approximation with
  Applications to Q-Learning}.
\newblock In \emph{International Conference on Artificial Intelligence and
  Statistics (AISTATS)}, volume 108 of \emph{Proceedings of Machine Learning
  Research}, pages 3015--3024. {PMLR}, 2020.

\bibitem[Zhang et~al.(2021)Zhang, Zhang, and
  Maguluri]{avg-td-Zhang2021FiniteSA}
S.~Zhang, Z.~Zhang, and S.~T. Maguluri.
\newblock Finite sample analysis of average-reward td learning and
  \$q\$-learning.
\newblock In \emph{NeurIPS}, 2021.

\end{thebibliography}

%





\end{document}